\documentclass[11pt]{article}

\usepackage{acl}

\usepackage{times}
\usepackage{latexsym}
\usepackage[T1]{fontenc}
\usepackage[utf8]{inputenc}
\usepackage{microtype}
\usepackage{inconsolata}
\usepackage{graphicx}
\usepackage{booktabs}
\usepackage{placeins}
\usepackage{multirow}
\usepackage{amsmath}
\usepackage{amssymb}
\usepackage{xspace}

\usepackage{xcolor}

\usepackage{comment}
\usepackage{tikz}
\usetikzlibrary{positioning, arrows.meta}
\usepackage{listings}
\usepackage[most]{tcolorbox}
\usepackage{pifont}
\usepackage{enumitem}

\usepackage[vlined,noend]{algorithm2e}
\usepackage{kotex}
\RestyleAlgo{boxruled}
\DontPrintSemicolon

\newcommand{\FrameworkName}{APT-RAG\xspace}
\newcommand{\FrameworkCall}{\operatorname{APT\mbox{-}RAG}}
\newcommand{\ours}{\FrameworkName}

\newcommand{\pstar}{Plan*RAG\xspace}
\newcommand{\toq}{ToQ\xspace}

\newcommand{\monaco}{MoNaCo\xspace}
\newcommand{\qampari}{QAMPARI\xspace}

\newcommand{\ans}{A}

\newcommand{\ques}{Q}
\newcommand{\quesik}[1]{\ques_{#1}}

\newcommand{\subqs}{q}

\newcommand{\quer}{\bar{Q}}
\newcommand{\querik}[1]{\quer_{#1}}

\newcommand{\set}[1]{\mathcal{#1}}

\newcommand{\corpus}{\set{D}}

\newcommand{\doci}[1]{d_{#1}}

\newcommand{\evi}{\set{E}}

\newcommand{\ncluster}{K}
\newcommand{\nclusterR}{\tilde{\ncluster}}
\newcommand{\nclusterA}{\hat{\ncluster}}
\newcommand{\optimal}[1]{{#1}^*}
\newcommand{\thsim}{\tau^{S}}
\newcommand{\thlen}{\tau^{L}}

\newcommand{\circled}[1]{%
  \tikz[baseline=(char.base)]{
    \node[shape=circle, draw, inner sep=0.5pt] (char) {\small #1};
  }%
}

\newcommand{\minisection}[1]{\vspace{-0.03in}\paragraph{#1}}

\title{A Tree-based RAG Framework for Evidence-Intensive QA via Adaptive Planning and Topology-Aware Evidence Gathering}

\author{
  Songeun Lee\textsuperscript{1}\thanks{Equal contribution.}
  \quad
  Kyungjin Min\textsuperscript{1,*} 
  \quad
  Injae Na\textsuperscript{2,*}
  \\
  \bfseries Suyeong Lee\textsuperscript{3}
  \quad
  Chiyoung Kim\textsuperscript{3}
  \quad
  Woohwan Jung\textsuperscript{1}\thanks{Corresponding author.}
  \\
  Korea University\textsuperscript{1}
  \quad
  Hanyang University\textsuperscript{2}
  \quad
  Hyundai Motor Company\textsuperscript{3}
  \\
  \texttt{\{lse173, kjmin\}@korea.ac.kr}
  \quad
  \texttt{suhoij47@hanyang.ac.kr}
  \\
  \texttt{\{lsy0620, cyoung.kim\}@hyundai.com}
  \quad
  \texttt{woohwan@korea.ac.kr}
}

\begin{document}
\maketitle

\begin{abstract}
Recent structured RAG methods leverage tree- or graph-based reasoning structures to improve multi-hop QA.
However, they face key limitations in evidence-intensive QA, where answering a question requires synthesizing information scattered across dozens or even hundreds of documents: structural rigidity, which limits adaptive reasoning expansion, and topology-ignorant evidence gathering, which prevents effective integration of evidence across different reasoning nodes.
To address these issues, we propose \FrameworkName{}, an \textbf{A}daptive \textbf{P}lanning and \textbf{T}opology-aware evidence gathering RAG framework. 
Adaptive planning dynamically expands the reasoning structure based on question dependencies and evidence requirements, while topology-aware evidence gathering improves evidence coverage through sibling evidence reuse, direct retrieval, and evidence aggregation from child nodes. 
We further introduce evidence-guided batched answer generation to reduce significant generation overhead in evidence-intensive QA. 
In the experiments on evidence-intensive QA benchmarks, \FrameworkName{} outperforms existing structured RAG methods.
Our code is available at \url{https://github.com/hyudsl/APT-RAG}.

\end{abstract}

\section{Introduction}
\label{sec:introduction}

Retrieval-Augmented Generation (RAG)-based reasoning approaches~\citep{IRCoT,SelfRAG} have achieved remarkable success in multi-hop QA through iterative retrieval and reasoning.
However, due to the linear reasoning path of these  approaches, they often suffer from error propagation and limited exploration space. 
To address these limitations, recent structured RAG frameworks~\citep{Tree-of-Question,Plan*RAG,RT-RAG,LogicRAG} utilize explicit reasoning structures—such as trees or graphs—to effectively model the dependencies of sub-questions.

\begin{figure}[t]
    \centering
    \includegraphics[width=\columnwidth]{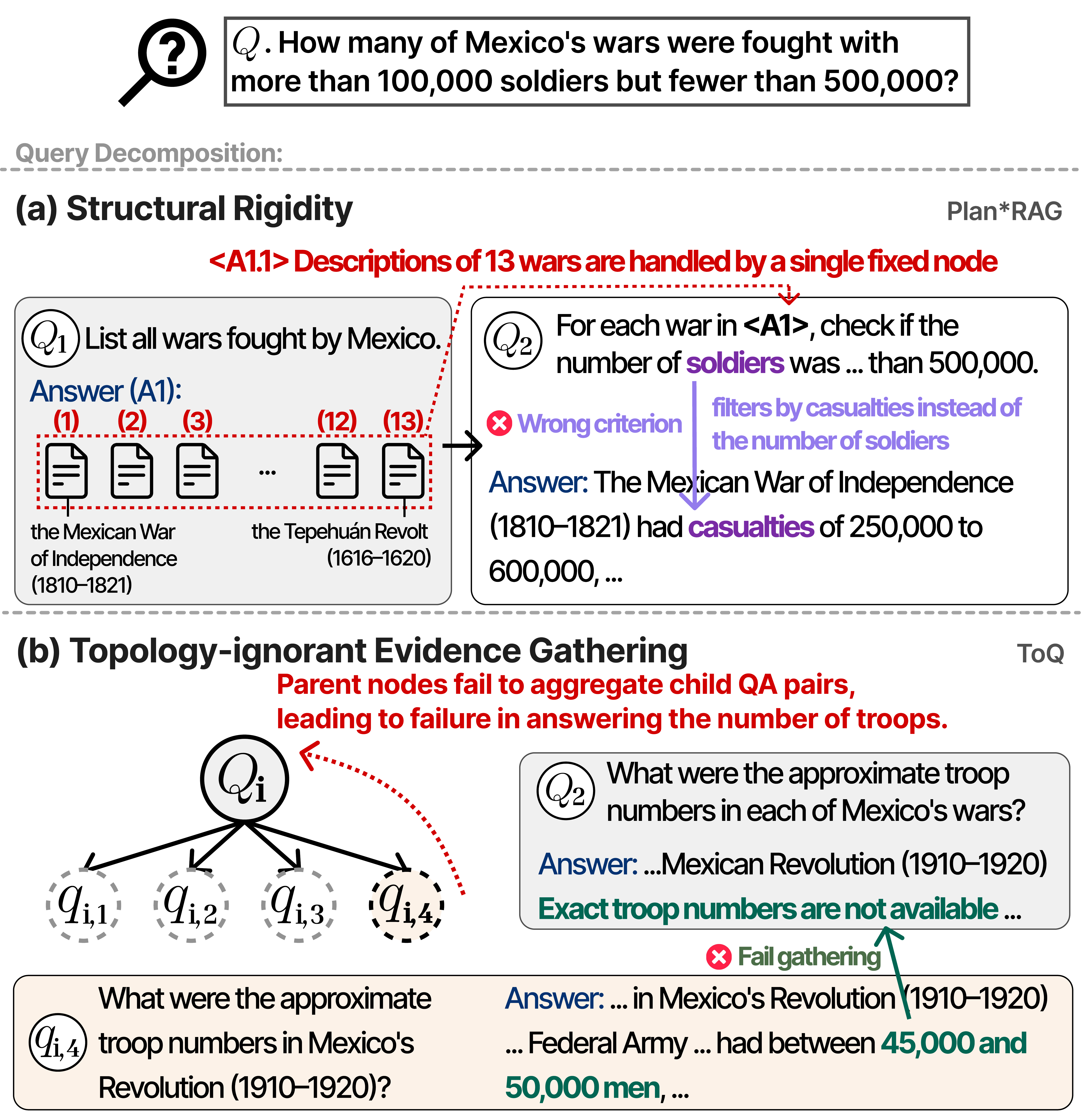}
    \caption{
    Failure cases of structured RAG frameworks caused by structural rigidity and topology-ignorant evidence gathering.
    }    
    \label{fig:structured_rag_limit}
\end{figure}

Despite their advantages, existing structured RAG frameworks face severe bottlenecks when transitioning to evidence-intensive QA~\citep{MoNaCo,QAMPARI}.
Unlike conventional multi-hop QA which requires only a few supporting pages on average (e.g., 2.0 pages in HotpotQA), 
evidence-intensive QA demands retrieving and synthesizing evidence distributed across dozens or hundreds of documents (e.g., 43.3 pages in MoNaCo).
This unprecedented scale of evidence exposes two fundamental structural deficiencies in current structured RAG methods.
First, existing methods, such as Plan*RAG\cite{Plan*RAG}, exhibit structural rigidity.
Since they construct a fixed reasoning plan upfront, they cannot dynamically construct the reasoning structure.
Consequently, a single node is often forced to process a lot of evidence when the reasoning becomes complex.
As illustrated in Figure~\ref{fig:structured_rag_limit}(a),
\pstar \cite{Plan*RAG} fails to answer question $Q_2$ since the node should synthesize 19 retrieved documents regarding 13 wars.
Second, other method such as \toq~\cite{Tree-of-Question} suffer from topology-ignorant evidence gathering.
Although these methods construct a reasoning architecture, they gather the evidence only in the root node and do not utilize the answers of the sub-questions in other nodes.
As shown in Figure~\ref{fig:structured_rag_limit}(b),
\toq fails to answer a sub-question $q_2$ even though it can be directly inferred from the answer of sub-question  $q_{2,4}$.

To address these limitations, we propose \FrameworkName (\textbf{A}daptive \textbf{P}lanning and \textbf{T}opology-aware evidence gathering) framework which answers evidence-intensive questions through a recursive depth-first process that constructs the reasoning tree.
At each step of traversal, the planner decides whether to further decompose a node, answer it directly, or resolve it using previously gathered evidence.
This planning allows the framework to flexibly adjust reasoning depth and breadth according to sub-problem complexity.
\ours also tightly couples tree traversal with hierarchical evidence gathering, preserving the reasoning topology through lateral gathering across sibling branches and vertical gathering from child to parent nodes.
This topology-aware synthesis enables each node to leverage relevant evidence for answer generation within its reasoning context.
Furthermore, \ours incorporates evidence-guided answer batching to reduce redundant generation over overlapping evidence, substantially lowering LLM calls.
On the evidence-intensive QA benchmarks \monaco~\citep{MoNaCo} and \qampari~\citep{QAMPARI}, \ours achieves higher answer F1 scores than existing structured RAG methods.
Additionally, we observed that \ours adaptively expands its reasoning tree and retrieves more supporting evidence as requirements increase while baseline methods retrieve a nearly constant number of documents regardless of requirements.

Our contributions are summarized as follows:
\begin{itemize}[itemsep=1pt, topsep=4pt, leftmargin=*]
    \item We identify structural rigidity and topology-ignorant evidence gathering as two key limitations of existing structured RAG methods in evidence-intensive QA.
    \item We propose \ours, a recursive framework that adaptively expands the reasoning tree and gathers evidence along its topology.
    \item We introduce evidence-guided batched answer generation to reduce inference latency.
    \item    We demonstrate that \ours consistently outperforms structured RAG baselines on evidence-intensive QA benchmarks.
    
\end{itemize}
\section{Related Work}
\label{sec:related_work}

\minisection{Evidence-intensive QA.}
Evidence-intensive QA is a question answering task that requires substantially more evidence and reasoning steps than conventional multi-hop QA benchmarks, including HotpotQA and MuSiQue~\citep{HotpotQA, MuSiQue}. 
Recent evidence-intensive QA benchmarks such as MoNaCo\citep{MoNaCo}, QAMPARI\citep{QAMPARI}, and DeepSearchQA~\citep{DeepSearchQA} require collecting and synthesizing evidence distributed in dozens to hundreds of documents, making it difficult to retrieve sufficient evidence through a single retrieval step using only the initial input question.

\minisection{Structured RAG for Multi-Hop QA.}
Structured RAG improves multi-hop QA performance by decomposing complex questions into sub-questions and organizing the reasoning process using tree- or graph-based structures. 
Tree-based approaches, such as \citet{Tree-of-Question, RT-RAG, prunerag}, decompose questions into sub-questions, construct a tree-structured reasoning architecture, and perform reasoning over the resulting tree structure.
Graph-based approaches utilize Directed Acyclic Graphs to control retrieval and reasoning flows.
In particular, \citet{Plan*RAG} constructs a fixed reasoning graph based on atomic sub-questions, while \citet{LogicRAG} optimizes retrieval order using a logical dependency graph and topological sorting.
However, although these methods effectively structure retrieval and reasoning at the sub-question level, they do not sufficiently address the dynamic expansion and management of evidence throughout the reasoning process, nor the hierarchical and lateral integration and synthesis of evidence required for evidence-intensive QA.

\section{Method}
\label{sec:method}
We first formulate the evidence-intensive QA, then introduce \FrameworkName{}, and finally present a method to improve its inference efficiency.

\begin{figure*}[t]
    \centering
    \includegraphics[width=\textwidth]{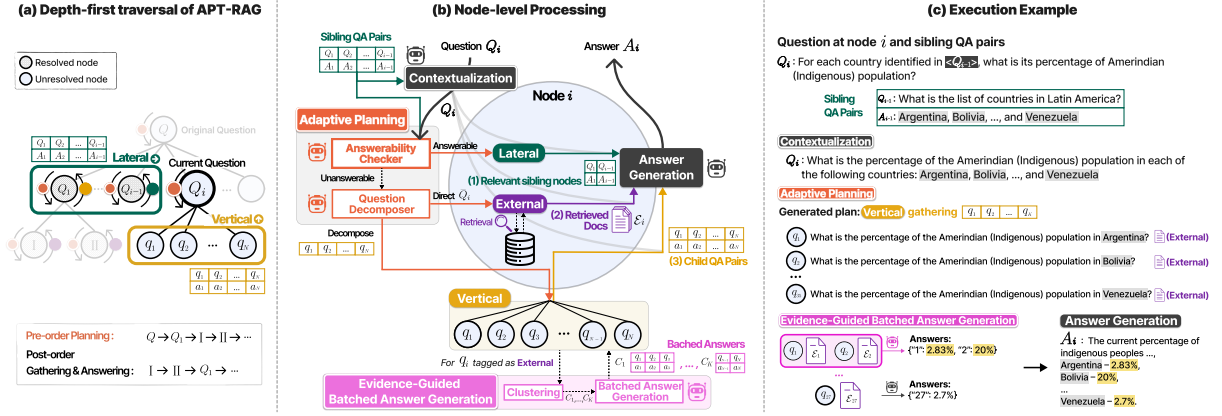}
    \caption{
    Overview of \textbf{\ours}.
    (a) Depth-first traversal with pre-order planning and post-order evidence gathering.
    (b) Node-level processing. (c) An example of node-level processing.
    }
    \label{fig:framework}
\end{figure*}

\subsection{Task Formulation}
\label{sec:task_formulation_notation}
Given a question $\ques$ and a corpus $\corpus = \{\doci{1}, ..., \doci{|\corpus|}\}$, the goal of evidence-intensive QA is to generate answer $\ans$ by synthesizing information distributed in dozens to hundreds of documents. 
Similar to conventional multi-hop QA, the question $\ques$ is decomposed into a set of sub-questions \(\mathcal{Q} = \{Q_1, Q_2, \dots, Q_N\}\), and a relevant evidence set $\evi_{Q_i} \subset \corpus$ is retrieved for each sub-question \(Q_i\).
However, unlike conventional multi-hop QA, where the required evidence is often limited to a small number of paragraphs, evidence-intensive QA requires substantially larger evidence sets. 
Consequently, the model must aggregate evidence and intermediate reasoning results across many reasoning steps to construct the final evidence set $\evi_Q$, from which the answer $\ans$ is generated.

\subsection{Overview of \FrameworkName{} Framework}
\label{sec:framework_overview}
Figure~\ref{fig:framework} shows an overview of \ours.
Given an original question $\ques$ and a corpus $\corpus$, 
\ours generates the final answer $\ans$ by recursively constructing and resolving a reasoning tree in a depth-first manner.
As shown in Figure~\ref{fig:framework}(a), \ours constructs a reasoning tree rooted at $\ques$ during execution, where each node represents a sub-question that should be resolved to answer its parent.
Rather than constructing a complete reasoning structure upfront, \ours interleaves planning, evidence gathering, and answer generation during recursive traversal.
This process is performed repeatedly on all nodes in the tree and the final answer $\ans$ to question $\ques$ is generated once the root node is resolved.
At each node $i$, \ours follows a four-step node-level procedure as shown in Figure~\ref{fig:framework}(b): contextualization, adaptive planning, topology-aware evidence gathering, and answer generation.

\minisection{Contextualization.} 
Following the previous structured RAG frameworks \cite{Plan*RAG,LogicRAG,RT-RAG, Tree-of-Question}, \ours can generate a sub-question $\quesik{i}$ that depends on the answers of preceding sibling nodes when the later sub-question is intended to use entities, values, or constraints identified by earlier ones.
We represent the dependency with a place holder with question IDs such as $<Q_{j}>$.
In the contextualization step, we resolve the reference by using the input questions with sibling question-answer pairs.
For example, in Figure~\ref{fig:framework}(c), question $\quesik{2}$ includes a reference to the answer of $\quesik{1}$, which contains a list of countries.
Contextualizer produces a self-contained question based on the given sibling pair $(\quesik{1}, \ans_1)$ by using a prompt in Appendix~\ref{app:prompts_contextualization}.
Note that if there is no dependency, the contextualization step does not change the input questions $\quesik{i}$.

\minisection{Adaptive Planning.} 
Next, \ours determines how evidence should be gathered to resolve the input question $\quesik{i}$. 
Depending on the dependency structure and evidence requirements of the question, the planner selects one of the following evidence gathering strategies:
\circled{1} reusing QA pairs from preceding sibling nodes,
\circled{2} directly retrieving evidence from the corpus, or
\circled{3} decomposing the question into sub-questions for fine-grained evidence exploration.

\minisection{Topology-aware Evidence Gathering.}
Based on the planning result, we gather evidence necessary to answer the question $\quesik{i}$ through one of three topology-aware strategies:
\circled{1} lateral gathering from preceding sibling QA pairs,
\circled{2} external gathering through direct retrieval, and
\circled{3} vertical gathering through aggregation of child QA pairs.

\minisection{Answer Generation.}
Finally, each node generates an answer $\ans_{i}$ to the input question $\ques_{i}$ based on the evidence $\evi_{Q_i}$. 
The generated answer can subsequently serve as evidence for sibling or parent nodes during later reasoning steps, allowing \ours to progressively accumulate and integrate information throughout the reasoning process. 
This recursive resolution process is repeatedly performed across the entire reasoning tree, ultimately producing the final answer $\ans$ to the question $\ques$.

\minisection{Scope of the Evidence.} Given a node $i$, every other node in the reasoning tree is in one of three regions: (i) its own subtree, (ii) its siblings' subtrees, or (iii) its ancestors and their siblings' subtrees.
\ours{} draws evidence from (i) and (ii)---via \emph{vertical} and \emph{lateral} gathering, respectively.
Vertical gathering propagates each child's answer along the tree topology rather than ingesting the full subtree at once, while lateral gathering reuses information already produced within the tree, preventing extra retrieval or decomposition at $i$.
Region (iii) is delegated to the ancestors of $i$, which incorporate this evidence during their own answer generation and thereby avoid redundant aggregation across the tree.
This topology-aware evidence flow allows each node to use only locally relevant evidence while avoiding redundant aggregation across the tree.

\subsection{Adaptive Planning}
\label{sec:planning}
Since the granularity and source of required evidence vary with the structural characteristics of each question, \ours adaptively determines an appropriate evidence gathering strategy during the planning step.
This stage consists of an answerability checker and a decomposer.

Given the input question $\quesik{i}$, the answerability checker first determines whether it can be resolved solely using QA pairs from previously processed sibling nodes.
If so, the lateral gathering is performed subsequently without further retrieval or decomposition, following the answerable path in Figure~\ref{fig:framework}(b).
Otherwise, $\quesik{i}$ is passed to the decomposer, which determines whether to keep the question as is or decompose it into sub-questions.
If the question is atomic and sufficient to search evidence through a single retrieval step, it is rewritten into a concise retrieval query $\querik{i}$; otherwise, it is decomposed into sub-questions, which are then organized as child nodes of $\quesik{i}$
The prompts for the planners are in Appendix~\ref{app:prompts_planning}

\subsection{Topology-aware Evidence Gathering}
\label{sec:evidence_gathering}

\minisection{Lateral Gathering.}
When question $\quesik{i}$ is determined to be answerable using preceding sibling QA pairs during the planning stage, \ours gathers sibling QA pairs relevant to $\quesik{i}$ as evidence. 
Such questions often depend on information already collected or generated from preceding nodes at the same depth, making them solvable without additional retrieval or decomposition.
Therefore, by reusing information obtained from preceding sibling nodes, lateral gathering effectively suppresses redundant tree expansion caused by repetitive retrieval and decomposition over already explored information~\citep{prunerag}.
This forms the first base case of the recursive construction.

\minisection{External Gathering.}
When the evidence required to answer $\quesik{i}$ can be directly obtained through retrieval without question decomposition, \ours performs cosine similarity-based semantic retrieval between embedding vectors as $\text{score}(\querik{i}, d_j) = 
\frac{\mathbf{e}(\querik{i})^\top \mathbf{e}(d_j)}{\|\mathbf{e}(\querik{i})\| \, \|\mathbf{e}(d_j)\|}$, using the retrieval query \(\querik{i}\) generated during the planning stage. 
The retrieved top-$k$ documents are then directly used as evidence for the input question.
This forms the second base case of the recursive construction.

\minisection{Vertical Gathering.}
Vertical gathering is performed when the question $\quesik{i}$ is decomposed into sub-questions. 
A single retrieval query may fail to sufficiently cover the broadly dispersed evidence required by complex questions. 
To address this, \ours decomposes the question into multiple sub-questions and gathers evidence at the child-node level, allowing each child node to focus on a narrower evidence scope. 
The QA pairs generated from the child nodes are then aggregated to expand the evidence coverage required to resolve the parent question.
This forms the recursive step of the construction.
When vertical gathering creates multiple independent child nodes that are planned for external gathering, \ours can further batch their answer generation using the evidence-guided clustering method described in Section~\ref{sec:evidence_aware_answer_batching}.

\begin{figure*}[tb]
    \centering
    \includegraphics[width=\textwidth]{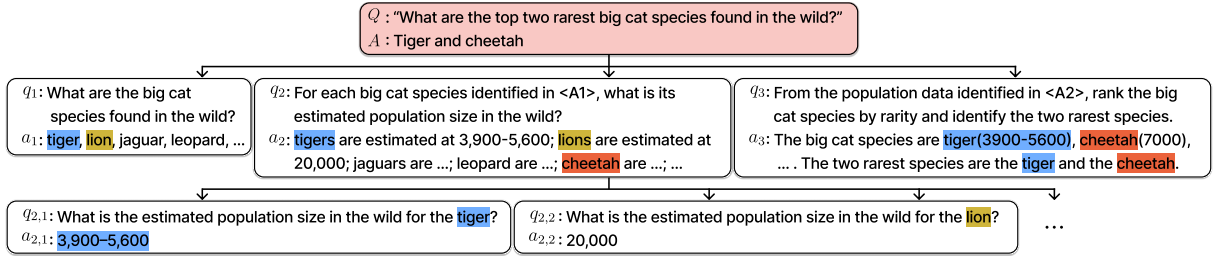}
    \caption{An example of the reasoning process of \ours}
    \label{fig:analysis_case_study}
\end{figure*}

\minisection{Example.}
Figure~\ref{fig:analysis_case_study} illustrates how the three gathering strategies operate within a single reasoning tree.
The initial question is first decomposed into multiple child nodes, where $q_1$ identifies candidate big cat species and $q_2$ is further decomposed into fine-grained sub-questions to retrieve population statistics for each species through external gathering.
The resulting lower-level QA pairs are aggregated through vertical gathering and reused as evidence for higher-level nodes.
Meanwhile, $q_3$ uses the QA pair generated at $q_2$ through lateral gathering to rank species by rarity, producing the final answer to the initial question.

\vspace{5 pt}
\begin{center}
    \includegraphics[width=\columnwidth]{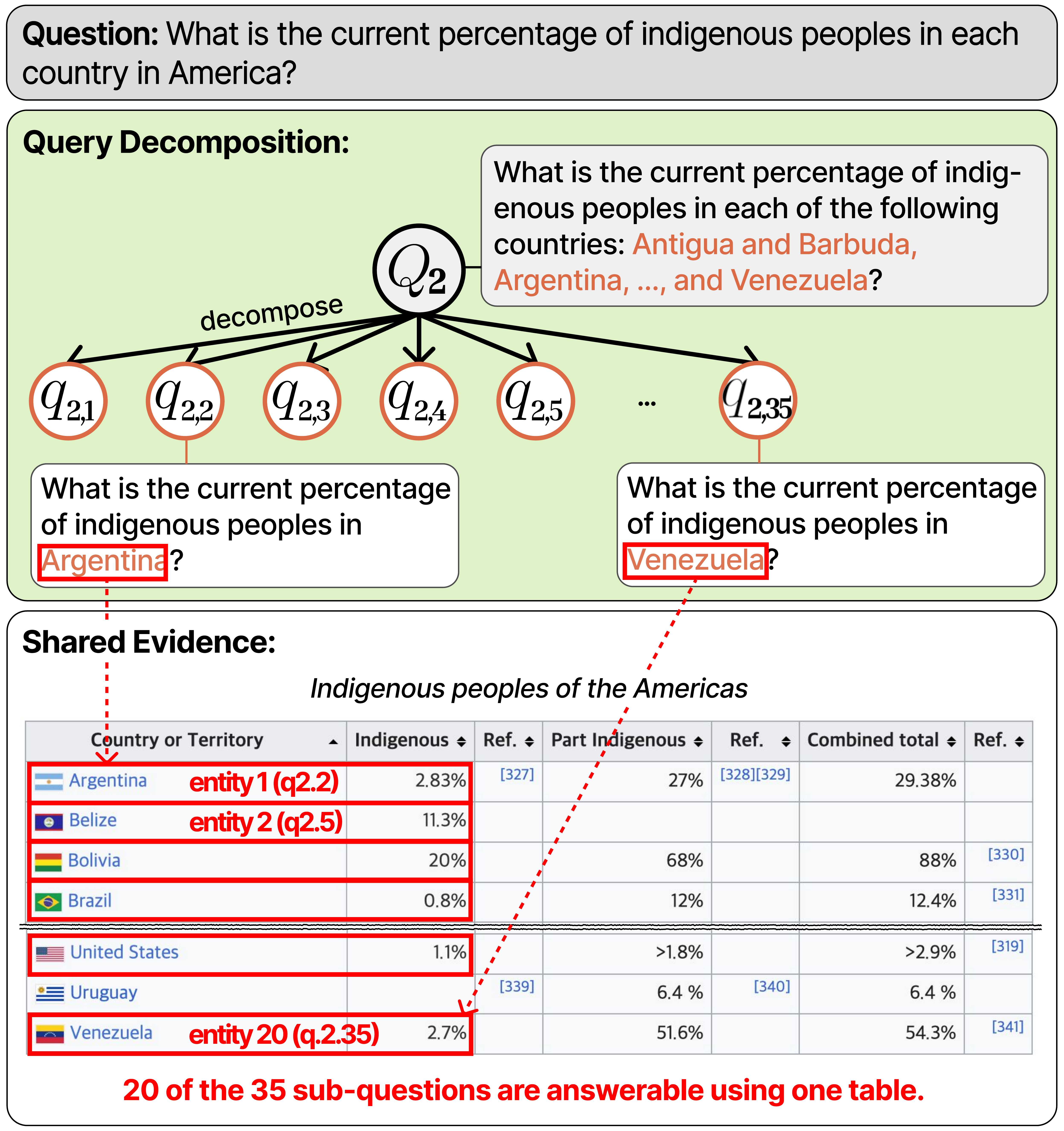}

    \captionof{figure}{
    An example of entity-specific sibling nodes from recall-oriented decomposition retrieving highly overlapping evidence from a shared Wikipedia page.
    }
    \label{fig:ec_motivation}
\end{center}
\vspace{3 pt}

\subsection{Evidence-Guided Batched Answer Generation}
\label{sec:evidence_aware_answer_batching}

When vertical gathering decomposes a node into multiple child sub-questions, many of the resulting sibling sub-questions often share the same evidence.
This often occurs when a parent question asks for the same attribute across multiple entities.
For example, as shown in Figure~\ref{fig:ec_motivation}, 
the parent question is decomposed into 35 sub-questions asking about the percent of indigenous peoples in 35 different the Americas.
Among them, 20 sub-questions can be answered by using a single shared evidence.
In that case, instead of independently answering each sub-question, 
\ours jointly answers sub-questions that share highly overlapping evidence.
It substantially reduces the number of LLM calls as well as the number of input tokens, thereby lowering latency for evidence intensive questions that require broad decomposition.

To do this, we slightly modify our pre-order planning schedule for the optimization.
When \ours decomposes a node into multiple child sub-questions, it first performs adaptive planning for all child nodes before starting evidence gathering for first child node. 
It then identifies the child nodes that are independent of preceding sibling answers and assigned to external gathering. 
These nodes can be answered jointly because they do not rely on one another and use only externally retrieved evidence. 
We cluster these candidate sub-questions based on the evidence overlap and answer the questions together with the union of their retrieved evidence.
The prompt for the batched answer generation is provided in Appendix~\ref{app:prompts_answering}.
In the rest of this section, we first formulate the sub-question clustering problem and then propose an approximate algorithm by reducing a relaxed clustering problem to the minimum clique partition problem.

%\memo{
%\minisection{Temp - clustering notation}
%Set of subquestions $\set{B} = \{q_1, \dots, q_n\}$ \\
%Number of clusters:\\
%~Original formulation:  $\ncluster$  (Optimal $\optimal{\ncluster}$)\\
%~Relaxed formulation: $\nclusterR$ (Optimal $\optimal{\nclusterR}$)\\
%~Solution of modified algorithm: $\nclusterA$\\ (Optimal $\optimal{\nclusterA}$)
%$\thlen$
%$\thsim$
%$\mathcal{D}$}
% \mathcal{Q} 수식 정의 해도 되는지 확인

\minisection{Problem Formulation.}
Let $\set{B} = \{q_1, \dots, q_n\}$ be the set of sub-questions selected for batching where each $q_i \in \set{B}$ has a set of retrieved documents $\mathcal{E}_i$.
We aim to partition $\set{B}$ into the minimum number of clusters $\{C_k\}_{k=1}^{\ncluster}$ while satisfying a context length constraint and an evidence-similarity constraint.

First, we limit the length of the batched prompt of each cluster by the following length constraint:
$\ell\left(\ \bigcup_{i \in C_k}\mathcal{E}_i\right) < \thlen $
where $\ell(\cdot)$ denotes token length and $L$ is the maximum evidence length allowed in a batch.
Second, sub-questions in the same cluster should have sufficiently overlapping evidence.
We measure the evidence similarity between two sub-questions $q_i$ and $q_j$ using the Jaccard similarity of their evidence sets: $s_{i,j}= \frac{|\mathcal{E}_i \cap \mathcal{E}_j|}{|\mathcal{E}_i \cup \mathcal{E}_j|}$
Accordingly, when sub-questions are grouped into a cluster $C_k$, the evidence fed to the language model is the deduplicated union  $\bigcup_{i \in C_k} \mathcal{E}_i$.
Greater overlap produces a more compact union, allowing more sub-questions to be handled per call while reducing redundant evidence processing.
By putting these together, the constrained sub-question clustering problem is formulated as:
\begin{subequations}
\begin{align}
&\text{minimize} ~ \ncluster\\
&\text{subject to}\quad\\
&\quad \bigcup_{k=1}^{\ncluster} C_k = \mathcal{B}, 
C_i \cap C_j = \emptyset (\forall i \ne j), \label{problem:eq:partition} \\
&\quad s_{ij} > \thsim  \quad \forall\, i, j \in C_k, \forall\, k\\
&\quad \ell\left(\ \bigcup_{i \in C_k}\mathcal{E}_i\right) < \thlen ~ \forall\, k
\end{align}
\end{subequations}
where the constraint in Eq.~\eqref{problem:eq:partition} ensures every sub-question belongs to exactly one cluster.

\minisection{Relaxation, Reduction, and Approximation.}
We reduce the constrained sub-question clustering problem to the \textit{Clique Covering Problem (CCP)}~\citep{ccp2013}, which seeks a minimum clique partition of the graph, by relaxing the context-window constraint.
Since the goal of evidence-guided clustering is to reduce LLM inference latency, the clustering step itself should be lightweight.
We therefore further reduce CCP to a graph coloring problem on the complement graph $\bar{G}$, and solve it with a modified greedy graph coloring algorithm with largest-first ordering~\citep{Welsh1967upper}.
The modification extends the original verification to additionally enforce the context-window constraint
$\ell\!\left(\bigcup_{i \in C_k \cup \{v\}} \mathcal{E}_i \right) < \thlen$
which yields a greedy solution $\nclusterA$ satisfying $\optimal{\nclusterR} \leq \optimal{\nclusterA} \leq \nclusterA$.
Details of the approximation algorithm and proofs for the problem reduction are provided in Appendix~\ref{app:theory}.

\minisection{Batched Answer Generation.}
For each cluster $C_k$, \FrameworkName{} concatenates the corresponding questions and the deduplicated documents into a single prompt and jointly generates answers:
\begin{equation*}
\{a_i\}_{i \in C_k}
=
\mathrm{Generator}\!\left(
\{ \subqs_i \}_{i \in C_k }
\;\|\;
\bigcup_{i \in C_k} \mathcal{E}_i
\right).
\end{equation*}

This module does not modify the reasoning tree structure or evidence semantics.
Instead, it improves inference efficiency by reducing redundant generation calls among sibling leaves that share overlapping retrieved evidence.

\begin{center}
\small
\begin{tabular}{l|cc}
\toprule
Statistic & MoNaCo & QAMPARI \\
\midrule
Num. Questions & 1,315 & 1,000 \\
Avg. Question Length (Chars) & 84.3 & 55.8 \\
Avg. Answer Length (Chars) & 241.2 & 274.3 \\
Avg. Gold Document & 43.3 & 13.0 \\
Size. Corpus (Document) & 784,776 & 6,311,809 \\
Size. Corpus (Chunks) & 1,212,878 & 25,856,230 \\
\bottomrule
\end{tabular}
\captionof{table}{Data statistics.}
\label{tab:dataset_stats}
\end{center}

\begin{table*}[t]
\centering
\small
\setlength{\tabcolsep}{3.2pt}
\resizebox{\textwidth}{!}{%
\begin{tabular}{clcccccccc}
\toprule
\multirow{2}{*}{Models} & \multicolumn{1}{c}{\multirow{2}{*}{Methods}} & \multicolumn{4}{c}{MoNaCo} & \multicolumn{4}{c}{QAMPARI} \\
\cmidrule(lr){3-6} \cmidrule(lr){7-10}
 &  & Ans. P & Ans. R & Ans. F1 & Ret. R & Ans. P & Ans. R & Ans. F1 & Ret. R \\
\midrule
\multirow{6}{*}{\textsc{Qwen3-4B-Inst}} & LLM-Only & 31.19 & 28.47 & 28.08 & -- & 9.32 & 4.25 & 4.97 & -- \\
 & NaiveRAG & 40.36 & 34.53 & 34.98 & 31.68 & \underline{31.65} & 16.44 & 18.58 & 11.74 \\
 & Plan$^\ast$RAG & 30.88 & 26.55 & 27.02 & 31.96 & 12.41 & 4.19 & 5.40 & 5.70 \\
 & LogicRAG & \underline{43.83} & 34.09 & 35.75 & 27.11 & 29.45 & 7.05 & 9.87 & 10.23 \\
 & \toq & 41.31 & \underline{36.22} & \underline{36.32} & \underline{33.07} & 31.18 & \underline{16.78} & \underline{18.76} & \underline{11.96} \\
 & \FrameworkName{} (Ours) & \textbf{46.75} & \textbf{39.95} & \textbf{40.69} & \textbf{40.12} & \textbf{34.12} & \textbf{18.78} & \textbf{20.91} & \textbf{14.08} \\
\midrule
\multirow{6}{*}{\textsc{Qwen3-30B-Inst}} & LLM-Only & 46.43 & 41.92 & 41.82 & -- & 19.76 & 7.35 & 9.40 & -- \\
 & NaiveRAG & 51.98 & \underline{46.56} & 46.64 & 31.68 & 33.43 & 18.46 & 20.46 & 11.74 \\
 & Plan$^\ast$RAG & 52.40 & 46.54 & \underline{47.03} & \underline{39.28} & 33.08 & 15.31 & 17.67 & \underline{16.54} \\
 & LogicRAG & \textbf{55.70} & 44.62 & 46.41 & 21.64 & \textbf{34.60} & 7.90 & 11.23 & 8.47 \\
 & \toq & 52.19 & 46.51 & 46.66 & 31.91 & 33.85 & \underline{19.66} & \underline{21.37} & 11.42 \\
 & \FrameworkName{} (Ours) & \underline{55.32} & \textbf{50.75} & \textbf{50.84} & \textbf{50.79} & \underline{34.51} & \textbf{22.08} & \textbf{23.28} & \textbf{19.61} \\
\bottomrule
\end{tabular}%
}
\caption{Performance comparison on evidence-intensive QA benchmarks.
}
% Bold means best results ans underlined means second-best.
\label{tab:main_results}
\end{table*}

\section{Experiments}
\label{sec:experiments}

\subsection{Experimental Setup}
\label{sec:experimental_setup}

\minisection{Datasets.}
We evaluate \FrameworkName{} on evidence-intensive QA benchmarks with different evidence scales:
(1) MoNaCo~\citep{MoNaCo}, a benchmark of complex information-seeking questions requiring synthesis across dozens to hundreds of evidence pages (43.3 pages on average); and
(2) QAMPARI~\citep{QAMPARI}, a multi-answer open-domain QA benchmark where answers are distributed across multiple passages, requiring aggregation over 13.0 evidence pages on average.
Statistics of the datasets are summarized in Table~\ref{tab:dataset_stats}.
Dataset examples are presented in Appendix~\ref{app:datasets}, and detailed corpus construction and indexing are described in Appendix~\ref{app:corpus}.

\minisection{Baselines.} 
We compare our method with two categories of baselines:
(1) basic baselines, including \textbf{LLM-Only} and \textbf{Naïve RAG}; and
(2) structured RAG baselines, including tree-based approaches such as \textbf{ToQ}~\citep{Tree-of-Question} and \textbf{RT-RAG}~\citep{RT-RAG}, and graph-based approaches such as \textbf{Plan*RAG}~\citep{Plan*RAG} and \textbf{LogicRAG}~\citep{LogicRAG}.
Due to the high cost of inference RT-RAG, we selected a representative 300-example subset of each benchmark for evaluating to keep computational costs manageable. We report the results in Appendix~\ref{app:subset}. 

\minisection{Evaluation Measures.}
We evaluate answer quality using macro-averaged answer precision, recall, and F1, following the evaluation protocols of \monaco and \qampari.
We also report retrieval recall(Ret. R), defined as the fraction of retrieved gold documents, to measure evidence coverage.
% LDfQ 1, KnUc 5 - Inference cost
We further evaluate inference cost by measuring the number of LLM calls, input and output tokens, and end-to-end latency per question, as detailed in Appendix~\ref{app:full_eval}.

\minisection{Implementation details.}
To ensure fair comparison, all methods use the same retrieval setup and backbone LLMs.
We use Qwen3-Embedding-0.6B as the embedding model in the dense passage retrieval and retrieve the top $k=20$ passages for all experiments.
Please refer to Appendix~\ref{app:corpus} for the detailed corpus construction and indexing.
As backbone language models, we use Qwen3-4B\footnote{Qwen3-4B-Instruct-2507}  and Qwen3-30B\footnote{Qwen3-30B-A3B-Instruct-2507}.
% KnUc 4
For \monaco, which requires LLM-as-a-judge evaluation, we use GPT-5.4 as the evaluator for the main results. Gemini-3.5-Flash yields comparable results, which are reported in Appendix~\ref{app:evaluator_robustness}.
Unless otherwise specified, all subsequent analyses are conducted on \monaco using Qwen3-30B-Inst as the backbone model.
% 8E4G 1
For the evidence-guided clustering in EC, we set the similarity threshold to $\thsim=0.0$, requiring at least one overlapping passage between subquestions in the same cluster, to maximize the latency reduction.
We set the context-window budget to $\thlen=150{,}000$ tokens for Qwen3-4B-Inst and $\thlen=50{,}000$ tokens for Qwen3-30B-Inst.
All prompts used in our experiments are provided in Appendix~\ref{app:prompts}.
All experiments were conducted on an NVIDIA RTX PRO 6000 GPU.

\begin{figure}[t]
    \centering
    \includegraphics[width=\columnwidth]{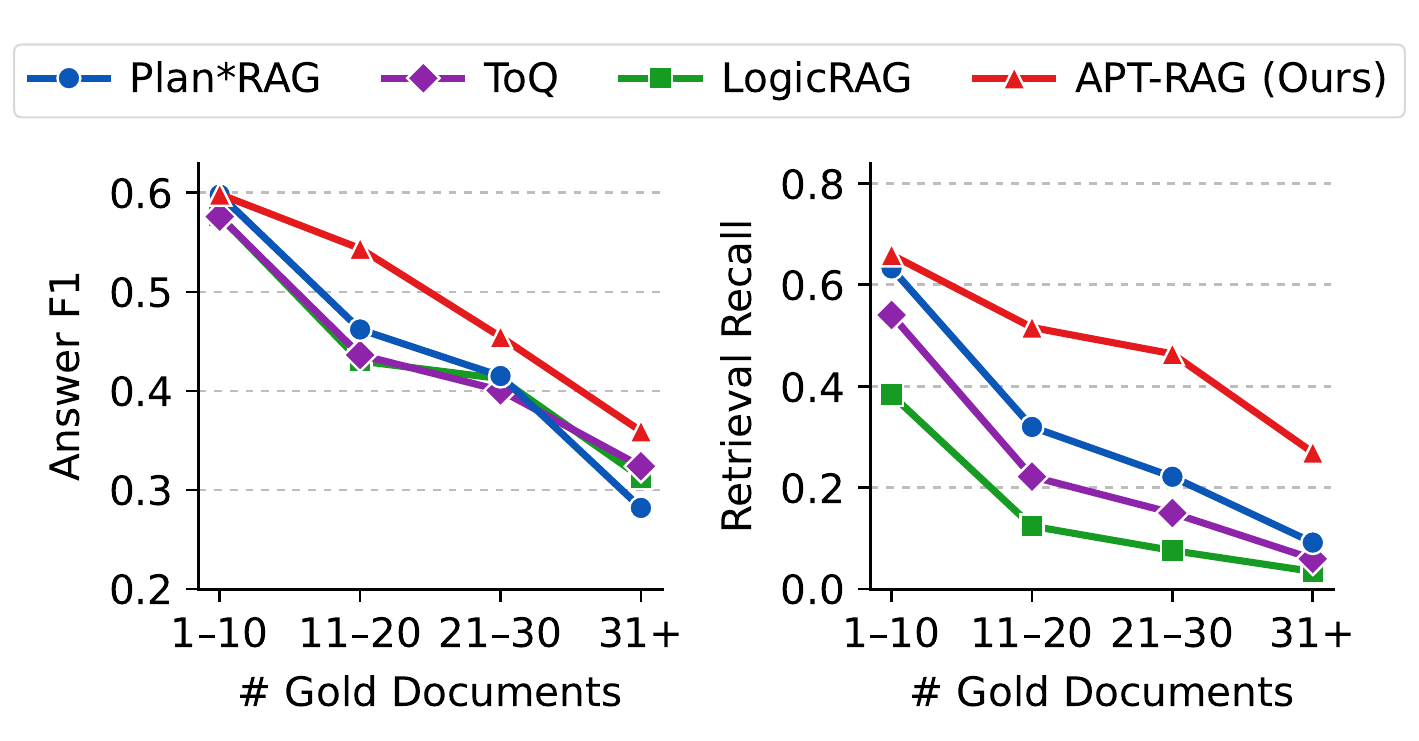}
    \caption{
    Comparison of answer F1 and retrieval recall across gold documents scale on MoNaCo ($\uparrow$ is better).}
    \label{fig:analysis_per_evidence}
\end{figure}

\subsection{Main Results}
\label{sec:main_results}
Table~\ref{tab:main_results} shows the answer quality and retrieval performance of \ours and the baselines. 
\ours achieves the best overall performance across all benchmarks and model scales, demonstrating its effectiveness for evidence-intensive QA.
All answer F1 gains of \ours over the baselines are statistically significant ($p \le 0.001$). Appendix~\ref{app:significance} reports each gain with its 95\% confidence interval.
On MoNaCo under the 30B setting, \ours outperforms the baseline, \pstar, by 8\% in answer F1, suggesting more effective evidence gathering for large-scale evidence synthesis. 
Specifically, \ours achieves a retrieval recall of 50.79, while \pstar reaches only 39.28. 
In contrast, although LogicRAG achieves the highest answer precision, its limited evidence coverage results in the lowest retrieval recall and ultimately the lowest answer F1 among non-LLM-only methods.
Under the 4B setting, most baselines show substantial degradation compared to their 30B counterparts, whereas \ours maintains relatively stable performance. 
A similar trend is observed on QAMPARI.

One key challenge in evidence-intensive QA is maintaining reasoning performance as the number of required evidence documents increases. 
Figure~\ref{fig:analysis_per_evidence} compares performance trends with respect to the number of required gold evidence pages. 
Although all methods degrade as evidence requirements grow, \ours consistently maintains the highest performance with a smaller drop than the baselines.
In particular, \ours achieves an answer F1 in the 21–30 page range comparable to the baselines in the 11–20 page range, while maintaining stable retrieval recall even in the 31+ range. 
These results suggest that \ours progressively expands the reasoning tree during inference, broadening retrieved evidence coverage and improving performance in evidence-intensive QA.

\FloatBarrier
\vspace{6 pt}
\begin{center}
    \includegraphics[width=\columnwidth]{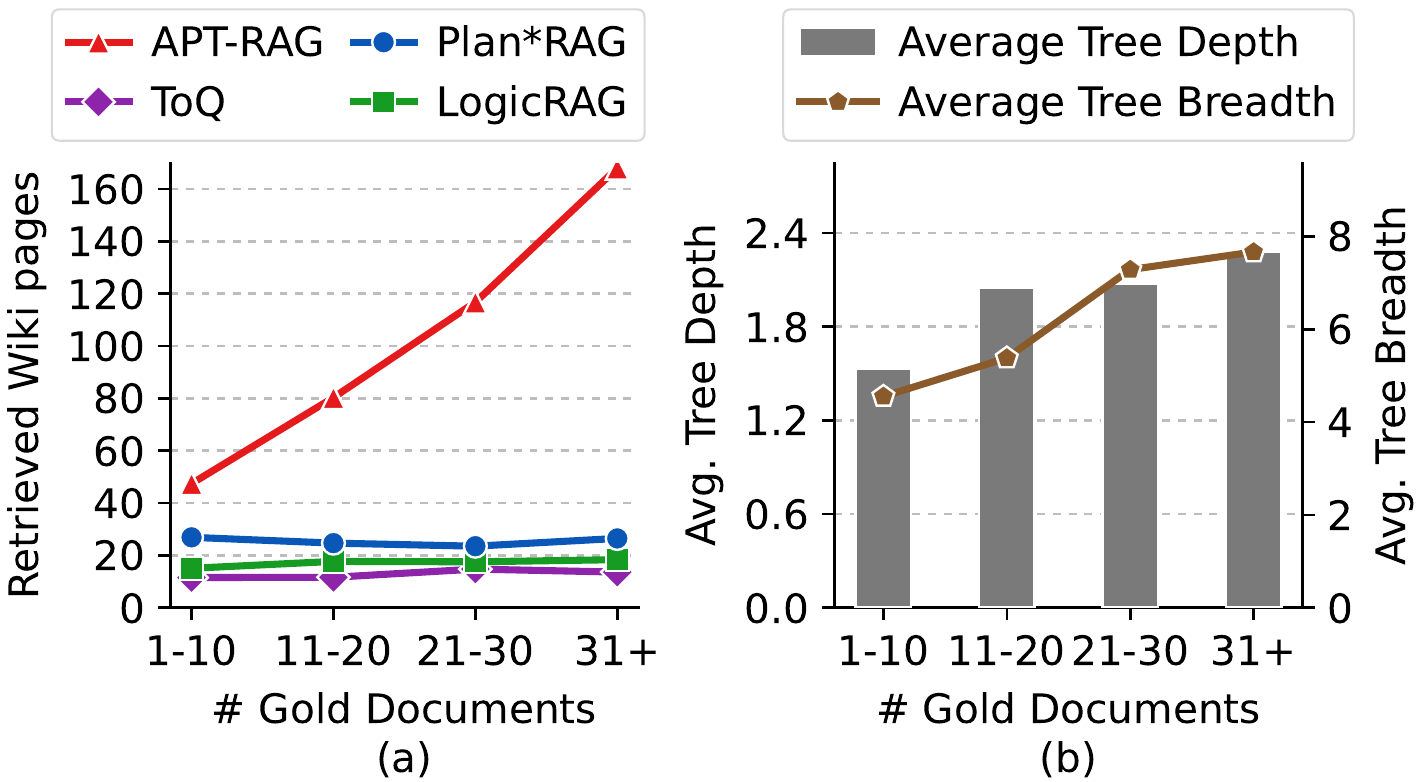}
    \captionof{figure}{
    Adaptive retrieval across the evidence scale on MoNaCo.
    (a) Number of retrieved documents for each method.
    (b) Average tree depth and branching factor of \FrameworkName{}.
    }
    \label{fig:analysis_depth_breadth_adaptation}
\end{center}

\subsection{Analysis}
\label{sec:analysis}

\minisection{Adaptive Retrieval Scaling.}
Figure~\ref{fig:analysis_depth_breadth_adaptation}(a) shows the number of retrieved documents with varying gold evidence scales. 
While all baselines retrieve a nearly constant number of documents regardless of the required evidence volume, \ours  increases the number of retrieved documents as the number of gold documents grows. 
This limitation stems from the structural rigidity of methods such as Plan*RAG and LogicRAG, whose fixed reasoning structures restrict further exploration, while ToQ prematurely terminates retrieval based on inaccurate sufficiency judgments (Appendix~\ref{app:baselines_methods}). 
Figure~\ref{fig:analysis_depth_breadth_adaptation}(b) further shows that the average tree depth and breadth of \ours increase together with the required evidence scale. 
In contrast, the baselines' depth and breadth remain nearly constant across the same range (Appendix~\ref{app:subset}).
It indicates that \ours adaptively expands both its retrieval space and reasoning structure according to query complexity.

\begin{table}[ht]
    \centering
    \small
    \begin{tabular*}{\columnwidth}{@{\extracolsep{\fill}}lccc@{}}
    \toprule
    Method & F1 $\uparrow$ & Ret. R $\uparrow$ & Latency $\downarrow$ \\
    \midrule
    \multicolumn{4}{@{}l}{w/o Performance-oriented modules} \\
    \quad w/o AP ($B=2$)
        & 49.61 & 37.48 & 46.56 \\
    \quad w/o AP ($B=\mathrm{Avg.}$)
        & 49.79 & \textbf{51.60} & 314.97 \\
    \quad w/o VG
        & 46.64 & 31.68 & 1.90 \\
    \midrule
    \multicolumn{4}{@{}l}{w/o Efficiency-oriented modules} \\
    \quad w/o EC \& LG
        & 53.96 & 53.29 & 183.95 \\
    \quad w/o LG
        & 52.24 & 54.56 & 178.55 \\
    \quad w/o EC
        & 52.34 & 50.69 & 113.90 \\
    \midrule
    \multicolumn{4}{@{}l}{Full framework} \\
    \quad APT-RAG (Ours)
        & \textbf{50.84} & 50.79 & \textbf{104.59} \\
    \bottomrule
    \end{tabular*}
    \caption{Component-wise ablation of APT-RAG. AP, VG, LG, and EC denote adaptive planning, vertical gathering,
    lateral gathering, and evidence-guided clustering, respectively.}
    \label{tab:component_ablation}
\end{table}
% \vspace{-6 pt}

\subsection{Ablation Studies}
\label{sec:Ablation Studies}

Table~\ref{tab:component_ablation} presents a component-wise ablation of \FrameworkName{}. 
We organize the ablations into two groups: performance-oriented modules, which affect evidence coverage and answer quality, and efficiency-oriented modules, which reduce inference cost.

% KnUc 6, ZSV5 3 - Ablation (method)
\minisection{Performance-oriented Modules.}
We evaluate adaptive planning by comparing \FrameworkName{} with fixed-breadth variants.
Fixing the tree breadth to 2 yields low latency but fails to capture varying evidence requirements across queries, resulting in the lowest answer F1 and retrieval recall. 
In contrast, fixing the tree breadth to the average breadth of 6 unnecessarily expands the tree for simple queries, substantially increasing latency.  % 8E4G 4 
Ours achieves the highest answer F1 and comparable retrieval recall to the average-breadth variant, while using less than half of its latency.
Additionally, removing vertical gathering, which restricts the framework to a single node and reduces it to NaiveRAG, decreases answer F1 by 4.20 points.
These results suggest that \ours effectively adapts the search structure to query-specific evidence requirements, balancing retrieval coverage and inference efficiency. 
Appendix~\ref{app:additional_ablation} further reports the same ablation under fine-grained evidence criterion.

\begin{center}
    \includegraphics[width=\columnwidth]{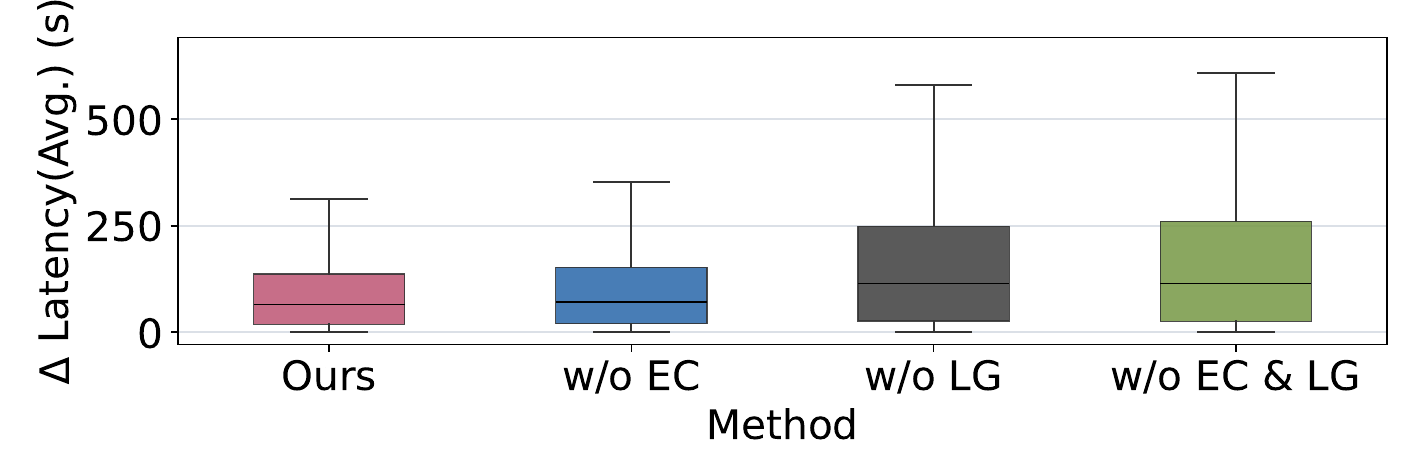}
    
    \captionof{figure}{
    Effectiveness of evidence-guided clustering (EC) and lateral evidence gathering (LG).
    }
    \label{fig:ablation_latency_distribution}
\end{center}

\begin{figure}[t]
    \centering
    \includegraphics[width=\columnwidth]{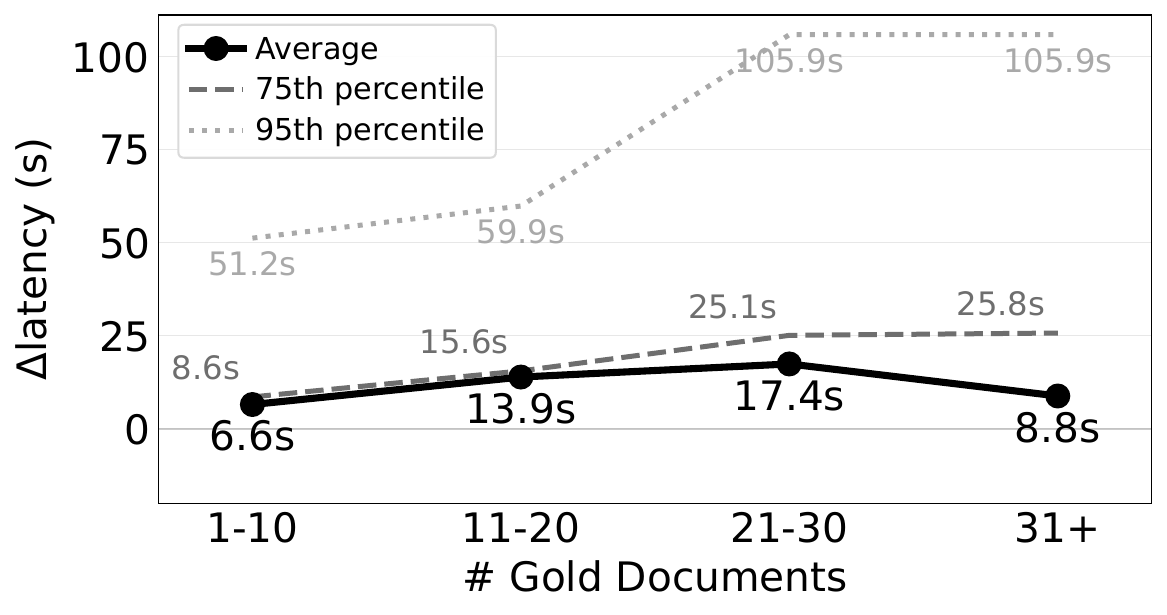}
\caption{
Effect of gold documents scale on latency reduction between \FrameworkName{} and \textit{w/o EC}. 
}
    \label{fig:ablation_latency_reduction}
\end{figure}

% KnUc 6, ZSV5 3 - Ablation (method)
\minisection{Efficiency-oriented Modules.}
We analyze the impact of lateral gathering and evidence-guided clustering on inference cost.
As shown in Table~\ref{tab:component_ablation}, lateral gathering and evidence-guided clustering reduce average inference latency by 41.4\% and 8.2\%, respectively.
For lateral gathering, Figure~\ref{fig:ablation_latency_distribution} shows that it controls worst-case latency by reducing redundant tree expansion caused by repetitive retrieval and decomposition over already explored information.
For evidence-guided clustering, Figure~\ref{fig:ablation_latency_reduction} shows its latency reduction across varying gold document scales.
While the average latency reduction increases relatively gradually,
\footnote{The average reduction decreases in the 31+ range because the larger merged evidence in this regime raises long-context processing cost, which outweighs the latency saved by clustering.} the latency reduction at the 75th and 95th percentiles becomes substantially larger as the number of gold documents increases.
In particular, at the 95th percentile, it achieves large latency reductions even for queries exhibiting extreme latency exceeding 100 seconds.
It shows that evidence-guided clustering effectively reduces long-tail inference latency.

\section{Conclusion}
\label{sec:conclusion}

We propose \ours, an adaptive planning and topology-aware evidence gathering RAG framework for evidence-intensive QA. 
By dynamically adjusting the search space and utilizing the reasoning topology, \ours overcomes the structural rigidity and topology-ignorant evidence gathering issues of existing structured RAG methods. 
Experiments show that \ours consistently improves answer quality and retrieval recall on evidence-intensive QA benchmarks.
These results suggest that recursive, topology-aware evidence gathering is a promising direction for answering questions over large, dispersed document spaces.

\section{Limitations}
\label{sec:limitations}

In this study, we focus on evidence-intensive QA tasks that require large-scale evidence aggregation, which can incur substantial inference cost and latency. 
Although evidence-aware batched answer generation reduces this overhead, the cost still increases as the amount of evidence and reasoning branches grows. 
This could be further mitigated through techniques such as tree pruning to control unnecessary branches and evidence expansion.

In addition, the performance of APT-RAG depends on the quality of the planning process that constructs the reasoning tree. 
In particular, inaccurate output from the query decomposer or answerability checker may generate unnecessary reasoning branches or propagate errors. 
Future work could further improve performance by training planning-related modules.

\section*{Acknowledgments}

This work was supported by Hyundai Motor Group through the 2025 Future Technology Research Program; the National Research Foundation of Korea (NRF) grant funded by the Korea government (MSIT) (RS-2025-16070246); the National R\&D Program for Cancer Control through the National Cancer Center (NCC) funded by the Ministry of Health \& Welfare, Republic of Korea (RS-2025-02264000); the IITP(Institute of Information \& Communications Technology Planning \& Evaluation)-ITRC(Information Technology Research Center) grant funded by the Korea government(Ministry of Science and ICT)(IITP-2026-RS-2024-00436857); the Institute of Information \& Communications Technology Planning \& Evaluation (IITP) grant funded by the Korea government (MSIT) (No. RS-2019-II190079, Artificial Intelligence Graduate School Program (Korea University)).

\raggedbottom
\bibliography{custom}

\appendix
% =====================================================================
% A. Datasets
% =====================================================================
\section{Datasets}
\label{app:datasets}

\begin{tcolorbox}[
    colback=white,
    colframe=black!80,
    arc=1mm,
    coltitle=white,
    colbacktitle=black!60,
    fonttitle=\bfseries,
    title={MoNaCo}
]
\small
\textbf{Question.} How did each of the Roman Emperors meet his end? How many years did they reign?

\textbf{Answer.}
\begin{itemize}[itemsep=6pt, parsep=0pt, topsep=6pt]
    \item `Basil I', `a fever contracted after a serious hunting accident when his belt was caught in the antlers of a deer...', `18'
    \item `Titus', `fever', `2'
    \item `Romanos III Argyros', `poison administered by his wife', `5'
    \item `Saloninus', `Postumus killed Saloninus', `0'
    \item \ldots
\end{itemize}
\end{tcolorbox}

\begin{tcolorbox}[
    colback=white,
    colframe=black!80,
    arc=1mm,
    coltitle=white,
    colbacktitle=black!60,
    fonttitle=\bfseries,
    title={QAMPARI}
]
\small
\textbf{Question.} What are the stage names of some famous drag queens?

\textbf{Answer.}
\begin{itemize}[itemsep=6pt, parsep=0pt, topsep=6pt]
    \item `Acid Betty'
    \item `Adore Delano'
    \item `Aja'
    \item `A\textquotesingle keria Chanel Davenport'
    \item `Akihiro Miwa'
    \item \ldots
\end{itemize}
\end{tcolorbox}

\begin{table*}[!t]
\centering
\small
\setlength{\tabcolsep}{6pt}
\begin{tabular}{llll}
\toprule
Method & Structure & Intermediate Synthesis Scope & Final Synthesis Scope \\
\midrule
\pstar & DAG & Retrieved docs & All nodes' QA pairs \\
LogicRAG & DAG & N/A & Accumulated summary \\
Tree-of-Question & Tree & Retrieved docs & Validated nodes' QA pairs \\
RT-RAG & Binary tree & Retrieved docs; child QA pairs & Root's child QA pairs \\
\textbf{\FrameworkName{}} (Ours) & Tree & Retrieved docs; sibling QA pairs; child QA pairs & Root's child QA pairs \\
\bottomrule
\end{tabular}
\caption{Structural comparison of evidence synthesis across structured RAG methods. \textit{Intermediate Synthesis Scope} lists the evidence each intermediate node consumes when generating its sub-answer; \textit{N/A} indicates that the method does not generate sub-answers. \textit{Final Synthesis Scope} lists the evidence aggregated for final answer generation.}
\label{tab:appendix_baseline_structure}
\end{table*}

% =====================================================================
% B. Corpus and Indexing
% =====================================================================
\section{Corpus and Indexing}
\label{app:corpus}

\paragraph{Corpus.}
We use English Wikipedia as the retrieval source for both benchmarks.
For QAMPARI, we adopt the official pre-chunked Wikipedia corpus released with the benchmark.
For MoNaCo, we construct the corpus by including all benchmark gold pages and sampling additional non-gold pages from the January 2026 English Wikipedia dump. 
We then chunk the selected pages using a rule-based chunker that preserves document structure: narrative text is split at sentence boundaries, list and infobox entries at line boundaries, and tables into row-level chunks with the header row retained.
Each passage is capped at 512 tokens using \texttt{tiktoken} and prepended with its article title and enclosing section heading, resulting in a collection of more than one million passages.

\paragraph{Indexing.}
We adopt a bi-encoder dense retriever~\citep{DPR}, using Qwen3-Embedding-0.6B without additional training.
All passage embeddings are indexed with faiss \texttt{IndexHNSWSQ} under SQ8 scalar quantization ($M{=}64$, $\text{efConstruction}{=}400$, $\text{efSearch}{=}128$, L2 distance), with the SQ8 quantizer trained on 100{,}000 sampled vectors. 
For MoNaCo, we additionally apply a section-level context expansion.
Upon retrieving a chunk, we collect additional chunks that share the same Wikipedia page, section heading, and content type (i.e., sentence, table, list, or infobox), and concatenate them with the retrieved chunk up to a maximum of 2{,}560 tokens.

% =====================================================================
% C. Baselines
% =====================================================================

\section{Baselines}
\label{app:baselines}

\subsection{Method Details}
\label{app:baselines_methods}

Table~\ref{tab:appendix_baseline_structure} compares the decomposition structure and evidence synthesis policy of each structured RAG baseline. 
The comparison describes the synthesis behavior used when the input question is decomposed into sub-questions; without decomposition, all methods generate the answer directly from retrieved documents.

\paragraph{\pstar.}
\pstar~\citep{Plan*RAG} represents the decomposition as a directed acyclic graph, where nodes are sub-questions and edges encode answer dependencies. 
Each intermediate node generates its sub-answer from retrieved documents. 
The final answer is generated from the QA pairs of all nodes in the graph.

\paragraph{LogicRAG.}
LogicRAG~\citep{LogicRAG} also uses a directed acyclic graph, but it does not generate a sub-answer for each node. 
It traverses nodes in topological order and incrementally refines a single information summary from the documents retrieved at each node. 
The final answer is generated from this accumulated summary.

\paragraph{Tree-of-Question.}
Tree-of-Question~\citep{Tree-of-Question} organizes sub-questions into a tree. 
Each node generates its sub-answer from retrieved documents, and an evaluator filters unreliable nodes during traversal. 
The final answer aggregates the QA pairs of all validated nodes.

\paragraph{RT-RAG.}
RT-RAG~\citep{RT-RAG} recursively splits each question into two children, forming a binary tree of bounded depth. 
Each intermediate node generates its answer from retrieved documents and child QA pairs. 
The final answer is generated from the QA pairs of the root's direct children. 
In its official setting, RT-RAG builds multiple candidate trees for each question, samples each tree several times to select a consensus structure, and repeatedly rewrites sub-questions during retrieval. 
This repeated tree construction, sampling, and rewriting increases the number of LLM calls, making RT-RAG slow in our evidence-intensive setting.

\subsection{Reproduction Details}
\label{app:baselines_reproduction}

We re-implement \pstar and \toq using the algorithms and prompts described in their original papers. 
% KnUc 1 - Unified final-answer prompt
To isolate the effect of the decomposition structure from the answer-generation policy, both use the \FrameworkName{} final-answer prompt (Appendix~\ref{app:prompts_answering}); it outperformed the original \pstar prompt, and \toq does not specify one.
We run LogicRAG\footnote{\url{https://github.com/chensyCN/LogicRAG}} and RT-RAG\footnote{\url{https://github.com/sakura20221/RT-RAG}} using the official implementations released by the authors. We keep their original reasoning algorithms and final-answer prompts.

% =====================================================================
% D. Algorithms
% =====================================================================
\begin{algorithm}[t]
\footnotesize
\caption{\FrameworkName without Batched Answer Generation}
\label{alg:agentic_rag_dynamic_planning}
\SetKwProg{Fn}{function}{:}{}
\Fn{$\FrameworkCall(Q_i, \mathcal{P}_i^{\mathrm{S}})$}{
    $Q_i \leftarrow \operatorname{Contextualizer}(Q_i, \mathcal{P}_i^{\mathrm{S}})$\;

    Evidence $\evi_{Q_i} \leftarrow \varnothing$\;

    \If{$\operatorname{AnswerabilityCheck}(Q_i, \mathcal{P}_i^{\mathrm{S}})$}{
        // \textit{Lateral gathering} \;
        $\evi_{Q_i} \leftarrow \mathcal{P}_i^{\mathrm{S}}$\;
    }  
    \Else{
         Subquestions $\mathcal{Q}_i \leftarrow \operatorname{Decomposer}(Q_i)$\;
        \eIf{$|\mathcal{Q}_i| = 0$}{
            // \textit{External gathering} \;
            $\evi_{Q_i} \leftarrow \operatorname{Retriever}(Q_i)$\;
        }{
            // \textit{Vertical gathering} \;
            \For{$k \leftarrow 1$ \KwTo $|\mathcal{Q}_i|$}{
                $a_k \leftarrow \FrameworkCall(q_k, \mathcal{P}_{i,<k}^{\mathrm{C}})$\;
            }
            Child QA Pairs $\mathcal{P}_i^{\mathrm{C}} \leftarrow \langle (q_k, a_k) \rangle_{k=1}^{|\mathcal{Q}_i|}$\;
            $\evi_{Q_i} \leftarrow \mathcal{P}_i^{\mathrm{C}}$\;
        }
    }

    // \textit{Answer generation} \;
    $A_i \leftarrow \operatorname{Generator}(Q_i, \evi_{Q_i})$\;
    \KwRet{$A_i$}\;
}
\end{algorithm}

\begin{algorithm}[ht]
  \footnotesize
  \caption{\FrameworkName with Batched Answer Generation}
  \label{alg:agentic_rag_dynamic_planning_t3}
  \SetKwProg{Fn}{function}{:}{}
  \Fn{$\FrameworkCall(Q_i, \mathcal{P}_i^{\mathrm{S}})$}{
      $Q_i \leftarrow \operatorname{Contextualizer}(Q_i, \mathcal{P}_i^{\mathrm{S}})$\;

      Evidence $\evi_{Q_i} \leftarrow \varnothing$\;
  
      \If{$\operatorname{AnswerabilityCheck}(Q_i, \mathcal{P}_i^{\mathrm{S}})$}{
          // \textit{Lateral gathering} \;
          $\evi_{Q_i} \leftarrow \mathcal{P}_i^{\mathrm{S}}$\;
      }  
      \Else{
           Subquestions $\mathcal{Q}_i \leftarrow \operatorname{Decomposer}(Q_i)$\;
          \eIf{$|\mathcal{Q}_i| = 0$}{
              // \textit{External gathering} \;
              $\evi_{Q_i} \leftarrow \operatorname{Retriever}(Q_i)$\;
          }{
              // \textit{Vertical gathering} \;
              \For{$k \leftarrow 1$ \KwTo $|\mathcal{Q}_i|$}{
                  $\rho_k \leftarrow \operatorname{PlanningMode}(q_k)$\;
              }
              $\mathcal{B} \leftarrow \varnothing$\;
              \For{$k \leftarrow 1$ \KwTo $|\mathcal{Q}_i|$}{
                  \eIf{$\rho_k = \textsc{External}$}{
                      $\mathcal{B} \leftarrow \mathcal{B} \cup \{k\}$\;
                  }{
                      $\mathcal{B} \leftarrow \operatorname{BatchedAnswer}(\mathcal{B})$\;
                      $a_k \leftarrow \FrameworkCall(q_k, \mathcal{P}_{i,<k}^{\mathrm{C}})$\;
                  }
              }
              $\mathcal{B} \leftarrow \operatorname{BatchedAnswer}(\mathcal{B})$\;
              Child QA Pairs $\mathcal{P}_i^{\mathrm{C}} \leftarrow \langle (q_k, a_k) \rangle_{k=1}^{|\mathcal{Q}_i|}$\;
              $\evi_{Q_i} \leftarrow \mathcal{P}_i^{\mathrm{C}}$\;
          }
      }
  
      // \textit{Answer generation} \;
      $A_i \leftarrow \operatorname{Generator}(Q_i, \evi_{Q_i})$\;
      \KwRet{$A_i$}\;
  }
  \BlankLine
  
  \Fn{$\operatorname{BatchedAnswer}(\mathcal{B})$}{
      \If{$\mathcal{B} \neq \varnothing$}{
          $\{\evi_{q_j}\}_{j \in \mathcal{B}} \leftarrow \operatorname{Retriever}(\{q_j\}_{j \in \mathcal{B}})$\;
          $\mathcal{C}_{\mathcal{B}} \leftarrow \operatorname{Cluster}(\{\evi_{q_j}\}_{j \in \mathcal{B}})$ \tcp*{Alg.~\ref{alg:greedy}}
          \ForEach{$C \in \mathcal{C}_{\mathcal{B}}$}{
              $\langle a_j \rangle_{j \in C} \leftarrow \operatorname{Generator}\bigl(\{q_j\}_{j \in C}, \textstyle\bigcup_{j \in C}\evi_{q_j}\bigr)$\;
          }
      }
      \KwRet{$\varnothing$}\;
  }
\end{algorithm}

\section{Algorithms}
\label{app:algorithm}

We present the pseudocode for the three algorithms used in \FrameworkName{}.

\subsection{Inference without Batched Answer Generation}
\label{app:inference_wo_batched}

% \paragraph{Inference without Batched Answer Generation.}
Algorithm~\ref{alg:agentic_rag_dynamic_planning} formalizes the recursive depth-first traversal introduced in \S\ref{sec:framework_overview}.
Each call $\FrameworkCall(Q_i, \mathcal{P}_i^{\mathrm{S}})$ resolves one node by executing the four-step node-level procedure: contextualization, adaptive planning (\S\ref{sec:planning}), topology-aware evidence gathering (\S\ref{sec:evidence_gathering}), and answer generation.
The planner directs the node to the lateral, external, or vertical branch, and the vertical branch expands the tree by recursing on each generated sub-question.
The algorithm uses four symbols that are specific to the pseudocode:
$\mathcal{P}_i^{\mathrm{S}}$ denotes the preceding sibling QA pairs supplied to node $i$,
$\mathcal{D}_i$ the documents returned by the retriever,
$\mathcal{P}_i^{\mathrm{C}}$ the child QA pairs aggregated from the recursive calls, and
$\mathcal{E}_i$ the evidence consumed by the answer generator, which is set to $\mathcal{P}_i^{\mathrm{S}}$, $\mathcal{D}_i$, or $\mathcal{P}_i^{\mathrm{C}}$ depending on the gathering mode.

\subsection{Inference with Batched Answer Generation}
\label{app:inference_w_batched}

Algorithm~\ref{alg:agentic_rag_dynamic_planning_t3} extends this recursive inference procedure with the batched answer generation described in \S\ref{sec:evidence_aware_answer_batching}.
After a node is decomposed into child sub-questions, it first computes $\rho_k \leftarrow \operatorname{PlanningMode}(q_k)$ for every child, where $\rho_k$ denotes the gathering mode planned for $q_k$.
The procedure then scans the children in order, accumulates consecutive children with $\rho_k=\textsc{External}$ into a batch $\mathcal{B}$, and flushes the batch before recursing on a child that requires non-external gathering.
Each flushed batch is handled by $\operatorname{BatchedAnswer}$, which retrieves evidence for the batched sub-questions, invokes $\operatorname{Cluster}$ procedure in Algorithm~\ref{alg:greedy} to group evidence-overlapping sub-questions under the context-window budget, and generates their answers with the union of the clustered evidence.
This preserves the original sibling-order dependencies for recursive children while jointly answering independent external children.

\begin{algorithm}[t]
\footnotesize
\caption{Constrained Greedy Graph Coloring with Largest-First Ordering}

% greedy graph-coloring based967 on the largest-first orderin
\label{alg:greedy}

\KwIn{
Sub-questions vertices $V=\{v_1,\ldots,v_n\}$,
retrieved evidence $\mathcal{S}=\{\mathcal{E}_1,\ldots,\mathcal{E}_n\}$
similarity threshold $\thsim$,
context budget $\thlen$
}
\KwOut{
Cluster set $\mathcal{C}=\{C_1,\ldots,C_{\nclusterA}\}$
}

\BlankLine

// \textit{Construct compatibility graph}

\ForEach{$i < j$}{
    $s_{ij}
    \leftarrow
    \operatorname{Jaccard}(\mathcal{E}_i,\mathcal{E}_j)$\;
    
    $\operatorname{compat}(i,j)
    \leftarrow
    \mathbf{1}[s_{ij} > \thsim]$\;
}

\BlankLine

// \textit{Largest-first ordering on complement graph $\bar{G}$}

\ForEach{$v_i \in V$}{
    $
    \deg_{\bar G}(v_i)
    \leftarrow
    |\{v_j : \neg \operatorname{compat}(i,j)\}|
    $\;
}

Sort $V$ in descending order of $\deg_{\bar G}$:
$
v_{\pi(1)},\ldots,v_{\pi(n)}
$\;

\BlankLine

\SetKwProg{Fn}{Function}{:}{}

\Fn{\textsc{IsValid}$(C_k, v)$}{
    
    \If{
        $\exists u \in C_k
        \text{ s.t. }
        \neg \operatorname{compat}(u,v)$
    }{
        \KwRet $\mathrm{False}$\;
    }

    \If{
        $
        \ell\!\left(
        \bigcup_{u \in C_k \cup \{v\}}
        \mathcal{E}_u
        \right)
        \ge \thlen
        $
    }{
        \KwRet $\mathrm{False}$\;
    }

    \KwRet $\mathrm{True}$\;
}

\BlankLine

// \textit{Greedy coloring}

$\mathcal{C} \leftarrow \varnothing$\;

\For{$i \leftarrow 1$ \KwTo $n$}{

    $\mathit{placed} \leftarrow \mathrm{False}$\;

    \ForEach{$C_k \in \mathcal{C}$}{
        
        \If{\textsc{IsValid}$(C_k, v_{\pi(i)})$}{

            $C_k \leftarrow C_k \cup \{v_{\pi(i)}\}$\;

            $\mathit{placed} \leftarrow \mathrm{True}$\;

            \textbf{break}\;
        }
    }

    \If{$\neg \mathit{placed}$}{
        $
        \mathcal{C}
        \leftarrow
        \mathcal{C}
        \cup
        \{\{v_{\pi(i)}\}\}
        $\;
    }
}

\KwRet{$\mathcal{C}$}

\end{algorithm}

\subsection{Evidence-Guided Subquestion Clustering}
\label{alg:ec}
Algorithm~\ref{alg:greedy} details the greedy approximation used in evidence-guided batched answer generation (\S\ref{sec:evidence_aware_answer_batching}).
In graph coloring, adjacent vertices must be assigned different colors.
We exploit this property by constructing the complement graph $\bar{G}$, where an edge connects two subquestion vertices that have \emph{insufficient} evidence overlap—precisely the pairs that should not share a cluster.
Applying graph coloring to $\bar{G}$ then ensures that incompatible subquestions are always assigned to different clusters.
Vertices are sorted in descending order of their degree in $\bar{G}$ (largest-first ordering), and a greedy coloring is applied in that order: each vertex is assigned to the first existing cluster that passes the \textsc{IsValid} check—requiring both pairwise evidence compatibility and the context-window budget $\thlen$—or placed into a new singleton cluster if no such cluster exists.
See Appendix~\ref{app:greedy} for a detailed description of the full algorithm.

% =====================================================================
% E. Evidence-Guided Batched Answer Generation 
% =====================================================================

\section{Clustering: Problem Reduction and Approximation Analysis}
\label{app:theory}

%% ----------------------------------------------------------------

\subsection{Constrained Clustering Formulation}
\label{app:constrained}

We formalize evidence-guided subquestion clustering as a constrained optimization problem.
Given a set of subquestions $\{q_i\}_{i=1}^n$ with retrieved evidence $\mathcal{E}_i$, our objective is to partition the subquestions into the minimum number of clusters while satisfying the context-window limit of the language model.

\paragraph{Graph Construction.}
To capture pairwise evidence compatibility, we construct a graph
\[
  G=(V, E),
\]
where each vertex corresponds to a subquestion and an edge indicates sufficient evidence overlap:
\begin{equation}
\begin{alignedat}{2}
  (i,j)\in E
  &\iff\;&
  \mathrm{Jaccard}(\mathcal{E}_i,\mathcal{E}_j) > \thsim ,
\end{alignedat}
\label{eq:graph_construction}
\end{equation}
for a similarity threshold $\thsim \ge 0$.

\paragraph{Constrained Clustering Objective.}
We seek the minimum number of clusters
\[
  \mathcal{C}=\{C_1,\dots,C_{\optimal{\ncluster}}\}
\]
such that each cluster contains mutually compatible subquestions and satisfies the context-window limit:
\begin{subequations}\label{eq:constrained_mcp}
\begin{align}
\optimal{\ncluster} = \min\;& \ncluster
\label{eq:constrained_mcp_obj}\\
\text{s.t.}\quad
&\bigcup_{k=1}^{\ncluster} C_k = V,
\label{eq:constrained_mcp_partition}\\
&C_a \cap C_b = \emptyset,
\quad (\forall a \ne b),
\label{eq:constrained_mcp_disjoint}\\
&\mathrm{Jaccard}(\mathcal{E}_i,\mathcal{E}_j) > \thsim,
\notag\\
&\qquad \forall i \ne j \in C_k,\ \forall k,
\label{eq:constrained_mcp_sim}\\
&\ell\!\left(
\bigcup_{i\in C_k}\mathcal{E}_i \right) < \thlen,
\quad \forall k.
\label{eq:constrained_mcp_ctx}
\end{align}
\end{subequations}
Here, $\ncluster$ denotes the number of clusters in a feasible solution, $\optimal{\ncluster}$ its minimum value, $L$ the context-window size limit, and $\ell(\cdot)$ the token count of the merged evidence.

\paragraph{Relaxation to Standard CCP.}
If constraint~\eqref{eq:constrained_mcp_ctx} is removed, the problem reduces to the standard clique covering problem (CCP) problem~\citep{ccp2013}:
\begin{equation}
\begin{aligned}
\optimal{\nclusterR}
&=
\min \; \nclusterR \\
\text{s.t.}\;
& \bigcup_{i=1}^{\nclusterR} C_i = V, \\
& C_a \cap C_b = \emptyset
\quad (\forall\, a \neq b), \\
& C_i \text{ induces a clique in } G
\quad (\forall\, 1 \le i \le \nclusterR).
\end{aligned}
\label{eq:mcp}
\end{equation}

The optimal value of this problem is the clique partition number of $G$, denoted by
\[
  \optimal{\nclusterR}=\theta(G).
\]

Since the constrained formulation introduces an additional context-window constraint, the feasible set of the constrained problem is a subset of that of the standard CCP.
Hence, the constrained optimum is at least as large:
\begin{equation}
  \optimal{\nclusterR} \;\le\; \optimal{\ncluster}.
  \label{eq:relax_ineq}
\end{equation}

%% ----------------------------------------------------------------

\subsection{Reduction to Graph Coloring}
\label{app:mcp:equivalence}

The CCP problem on $G$ is equivalent to graph coloring on its complement graph $\bar G=(V,\bar E)$, in which adjacent vertices are assigned different colors, where
\[
  (i,j)\in\bar E
  \iff
  (i,j)\notin E.
\]
In particular,
\begin{equation}
  \theta(G)=\chi(\bar G),
  \label{eq:mcp_gc_equiv}
\end{equation}
where $\theta(G)$ and $\chi(\bar G)$ denote the clique partition number of $G$ and the chromatic number of $\bar G$, respectively.
This equivalence follows because a set of vertices forms a clique in $G$ if and only if it forms an independent set in $\bar G$.
Therefore, partitioning $V$ into cliques of $G$ is equivalent to a proper vertex coloring of $\bar G$~\citep{West2001introduction}.

Under this equivalence, the compatibility condition in constraint~\eqref{eq:constrained_mcp_sim} corresponds to an independent-set condition on the complement graph
% \[
%   G
%   =
%   \bigl(
%     V,
%     \{(i,j):\mathrm{Jaccard}(\mathcal{E}_i,\mathcal{E}_j)\le S\}
%   \bigr).
% \]
\[
\bar{G} = (V, \{(i,j):\mathrm{Jaccard}(\mathcal{E}_i,\mathcal{E}_j)\le \thsim\})
\]
Therefore, our problem reduces to a graph coloring problem on $\bar{G}$
in which each color class must additionally satisfy the
context-window constraint
\[
  \ell\!\left(
    \bigcup_{i\in C_k}\mathcal{E}_i
  \right)
  < \thlen.
\]

%% ----------------------------------------------------------------

\subsection{Greedy Approximation}
\label{app:greedy}

% todo: NP-hard reference check
Since the constrained clustering problem remains NP-hard, we adopt a greedy graph-coloring based on the largest-first ordering~\citep{Welsh1967upper}.
The procedure is summarized in 
Appendix~\ref{alg:ec}.

\paragraph{Greedy approximation algorithm.}

Given the complement graph 
\[ \bar{G}=(V,\bar{E}),\] 
the coloring is represented by a function
\[
  f:V\to\{1,2,\dots,k\},
\]
where $f(v)$ denotes the color assigned to vertex $v$ and $k$ denotes the number of colors used.

\medskip

\textbf{[1 Initialization].}
Set $i=0$ and construct the complement graph
$\bar G=(V,\bar E)$ by adding an edge
$(u,v)\in\bar E$ whenever
\[
  \mathrm{Jaccard}(\mathcal E_u,\mathcal E_v)\le \thsim.
\]
Then compute the degree
\[
  \deg_{\bar G}(v)
  =
  |\{u\in V:(u,v)\in\bar E\}|
\]
for each vertex $v\in V$.
The vertices are then sorted in descending degree order as
\[
  \deg_{\bar G}(v_{\pi(1)})
  \ge
  \deg_{\bar G}(v_{\pi(2)})
  \ge
  \cdots
  \ge
  \deg_{\bar G}(v_{\pi(n)}),
\]
where $\pi$ denotes the resulting vertex ordering and ties are broken arbitrarily.
Finally, the color classes
\[
  C_1,C_2,\dots \subseteq V
\]
are initialized as empty sets.

\medskip

\textbf{[2 Next vertex].}
Increment $i$, and consider the next vertex $v_{\pi(i)}$.
If $i=n+1$, terminate with $f$ as the resulting coloring.

\medskip

\textbf{[3 Find admissible colors].}
Compute the set
\[
  F_i
  =
  \left\{
    f(v_{\pi(j)})
    :
    j<i,\,
    (v_{\pi(j)},v_{\pi(i)})\in\bar E
  \right\},
\]
that is, the set of colors already assigned to the neighbors of $v_{\pi(i)}$ in $\bar G$.

\medskip

\textbf{[4 Assign the smallest available color].}
Assign $v_{\pi(i)}$ to the first color class $C_r$ with
\[
  r \notin F_i
\]
satisfying
\[
  \ell\!\left(
    \bigcup_{u\in C_r\cup\{v_{\pi(i)}\}}
    \mathcal E_u
  \right)<\thlen.
\]
The vertex $v_{\pi(i)}$ is then assigned color $r$, i.e.,
\[
  f(v_{\pi(i)})=r.
\]
If no admissible color class satisfies the constraint,
a new singleton color class is created.

\medskip

Including the largest-first sorting step,
the overall running time of the greedy procedure is
\[
  O(m+n\log n),
\]
where $n=|V|$ and $m=|\bar E|$.

\vspace{12 pt}
\begin{center}
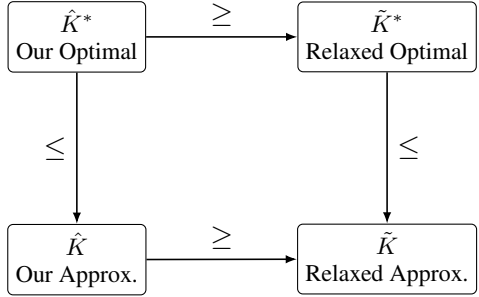

\begin{tikzpicture}[
    node distance=2.0cm and 2.0cm,
    box/.style={
        draw,
        rounded corners=2pt,
        inner sep=3pt,
        align=center,
        font=\small
    },
    arr/.style={-{Latex[length=1.5mm]}, semithick}
]
\node[box] (Kstar)  {$\optimal{\nclusterA}$\\{\footnotesize Our Optimal}};
\node[box, right=of Kstar]  (Kpstar) {$\optimal{\nclusterR}$\\{\footnotesize Relaxed Optimal}};
\node[box, below=of Kstar]  (K)      {$\nclusterA$\\{\footnotesize Our Approx.}};
\node[box, below=of Kpstar] (Kp)     {$\nclusterR$\\{\footnotesize Relaxed Approx.}};
\draw[arr] (Kstar) -- node[above] {$\geq$} (Kpstar);
\draw[arr] (K) -- node[above] {$\geq$} (Kp);
\draw[arr] (Kstar) -- node[left] {$\leq$} (K);
\draw[arr] (Kpstar) -- node[right] {$\leq$} (Kp);
\end{tikzpicture}

\captionof{figure}{
Relationship among optimal and approximate solutions
under the constrained and relaxed formulations.
}
\label{fig:reduction_structure}
\end{center}
\vspace{4 pt}

\subsection{Approximation Analysis}
\label{app:approx}

Let $\nclusterA$ and $\nclusterR$ denote the number of clusters produced by our greedy approximation algorithm and its relaxed version in Equation~\eqref{eq:mcp}, respectively, and let $\optimal{\nclusterA}$ and $\optimal{\nclusterR}$ denote their corresponding optimal values.
Figure~\ref{fig:reduction_structure} summarizes their relationships.

\paragraph{Approximation Hierarchy.}
\label{prop:hierarchy}
The following inequalities hold:
\begin{equation}
  \optimal{\nclusterR} \;\le\; \optimal{\nclusterA} \;\le\; \nclusterA, \qquad
  \optimal{\nclusterR} \;\le\; \nclusterR \;\le\; \nclusterA.
  \label{eq:hierarchy}
\end{equation}

\noindent
Each inequality follows directly:
\begin{itemize}
  \item $\optimal{\nclusterR} \le \optimal{\nclusterA}$: the relaxed problem admits more
        feasible solutions, so its optimal is smaller or equal.
\item $\optimal{\nclusterA} \le \nclusterA$: $\optimal{\nclusterA}$ is the optimal value,
      so $\nclusterA$ is an upper bound on it by definition.
\item $\optimal{\nclusterR} \le \nclusterR$: $\optimal{\nclusterR}$ is the optimal value,
      so $\nclusterR$ is an upper bound on it by definition.
\item $\nclusterR \le \nclusterA$: since the unconstrained greedy does not enforce
      the context-window constraint, it can merge clusters that
      the constrained greedy cannot, yielding fewer or equal clusters.
\end{itemize}

\noindent
Together, these inequalities confirm that ${\optimal{\nclusterR}}$ serves as a lower bound and $\nclusterA$ as an upper bound on the constrained optimum $\optimal{\nclusterA}$.

\begin{figure}[t]
\centering
\includegraphics[width=\linewidth]{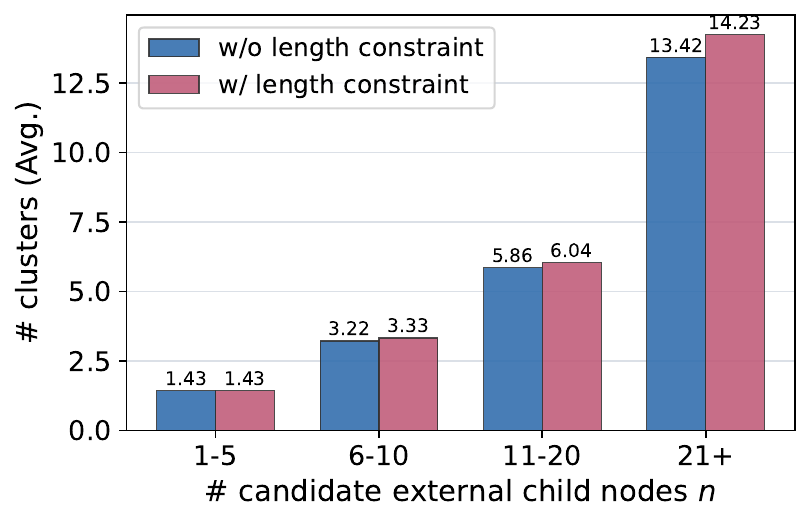}
\caption{
Effect of the context-window constraint on the number of clusters.
The x-axis denotes the number of candidate external child nodes $n=|B|$, while the y-axis shows the average number of clusters for the relaxed and constrained settings.
}
\label{fig:cluster_length_constraint}
\end{figure}

Figure~\ref{fig:cluster_length_constraint} evaluates the effect of introducing the context-window constraint into the greedy clustering algorithm.
As the number of candidate external child nodes increases, the difference between the relaxed and constrained settings remains small; even at the $21+$ point, the gap is less than one cluster.
The result suggests that the context-window constraint preserves clustering behavior close to the relaxed solution.

% =====================================================================
% F. Prompts
% =====================================================================
\section{Prompt Templates}
\label{app:prompts}

We provide the prompt templates used by the LLM-based components of \FrameworkName{} introduced in \S\ref{sec:method}.
The templates are organized according to the four stages of the node-level procedure: contextualization(Appendix~\ref{app:prompts_contextualization}), adaptive planning(Appendix~\ref{app:prompts_planning}), evidence gathering(Appendix~\ref{app:prompts_gathering}), and answer generation(Appendix~\ref{app:prompts_answering}).
For each component, we explain the role its prompt plays in the recursive traversal, how the instruction is structured to enforce that role, and the design decisions that motivated this structure.

\tcbset{
  appprompt/.style={
    colback=white,
    colframe=black!80,
    arc=1mm,
    boxrule=0.8pt,
    coltitle=white,
    colbacktitle=black!60,
    fonttitle=\bfseries,
    listing only,
    listing options={
      basicstyle=\ttfamily\scriptsize,
      breaklines=true,
      breakatwhitespace=true,
      columns=fullflexible,
      keepspaces=true
    }
  }
}

\subsection{Contextualization}
\label{app:prompts_contextualization}
Contextualization fills $<Q_{j}>$ in a sub-question using entities or values from the referenced sibling's answer (Figure~\ref{fig:prompt_contextualizer}; \S\ref{sec:framework_overview}).
The prompt uses a four-step sequence: analyze the question, identify needed conditions, extract items from prior answers, and then rewrite.
Listing the items in Step 3 before the rewrite in Step 4 avoids leaving the placeholder verbatim or vaguely paraphrased.

\begin{table*}[!t]
\centering
\setlength{\tabcolsep}{3.2pt}
\resizebox{\textwidth}{!}{%
\begin{tabular}{clccrrccrr}
\toprule
\multirow{2}{*}{Models} & \multicolumn{1}{c}{\multirow{2}{*}{Methods}} & \multicolumn{4}{c}{MoNaCo} & \multicolumn{4}{c}{QAMPARI} \\
\cmidrule(lr){3-6} \cmidrule(lr){7-10}
 &  & \# LLM Calls & \# Ret. Calls & Input Tok. & Output Tok. & \# LLM Calls & \# Ret. Calls & Input Tok. & Output Tok. \\
\midrule
\multirow{2}{*}{\textsc{Qwen3-4B-Inst}} & RT-RAG & 95.0 & 10.9 & 921,219 & 40,558 & 29.4 & 2.8 & 183,320 & 12,269 \\
 & \FrameworkName{} (Ours) & 16.4 & 7.1 & 69,982 & 1,699 & 4.9 & 1.7 & 8,366 & 227 \\
\midrule
\multirow{2}{*}{\textsc{Qwen3-30B-Inst}} & RT-RAG & 115.6 & 14.1 & 1,130,788 & 52,319 & 59.0 & 7.1 & 354,931 & 29,299 \\
 & \FrameworkName{} (Ours) & 32.1 & 13.5 & 134,923 & 4,238 & 11.5 & 4.3 & 21,447 & 1,072 \\
\bottomrule
\end{tabular}%
}
\caption{Per-example inference cost of RT-RAG and \FrameworkName{} on the 300-example subset of MoNaCo and QAMPARI. \# LLM Calls: average number of LLM invocations per question; \# Ret.\ Calls: retriever invocations; Input/Output Tok.: average input and output tokens per question.}
\label{tab:appendix_rt_cost}
\end{table*}

\begin{table*}[t]
\centering
\setlength{\tabcolsep}{3.2pt}
\resizebox{\textwidth}{!}{%
\begin{tabular}{clcccccccc}
\toprule
\multirow{2}{*}{Models} & \multicolumn{1}{c}{\multirow{2}{*}{Methods}} & \multicolumn{4}{c}{MoNaCo} & \multicolumn{4}{c}{QAMPARI} \\
\cmidrule(lr){3-6} \cmidrule(lr){7-10}
 &  & Ans. P & Ans. R & Ans. F1 & Ret. R & Ans. P & Ans. R & Ans. F1 & Ret. R \\
\midrule
\multirow{7}{*}{\textsc{Qwen3-4B-Inst}} & LLM-Only & 31.08 & 28.59 & 28.07 & -- & 12.73 & 5.38 & 6.56 & -- \\
 & NaiveRAG & 42.68 & 35.90 & 37.10 & 32.26 & 36.19 & 18.77 & 21.37 & 12.13 \\
 & Plan$^\ast$RAG & 31.77 & 28.61 & 29.14 & 33.52 & 16.60 & 5.94 & 7.52 & 6.87 \\
 & LogicRAG & 43.11 & 33.98 & 35.90 & 28.33 & 33.95 & 8.87 & 12.10 & 10.32 \\
 & RT-RAG & \textbf{46.78} & 38.13 & \textbf{39.86} & 21.90 & 37.90 & 14.93 & 17.86 & 13.21 \\
 & ToQ & 38.58 & 33.39 & 34.17 & 34.03 & 35.21 & 19.78 & 21.75 & 12.57 \\
 & APT-RAG (Ours) & 44.16 & \textbf{38.27} & 39.14 & \textbf{41.23} & \textbf{38.32} & \textbf{22.05} & \textbf{24.44} & \textbf{14.99} \\
\midrule
\multirow{7}{*}{\textsc{Qwen3-30B-Inst}} & LLM-Only & 44.22 & 40.35 & 39.94 & -- & 22.97 & 9.60 & 11.74 & -- \\
 & NaiveRAG & 54.33 & 47.70 & 48.27 & 32.26 & 38.14 & 20.72 & 23.19 & 12.13 \\
 & Plan$^\ast$RAG & 53.25 & 48.09 & 48.59 & 40.71 & 34.09 & 15.80 & 18.21 & 17.00 \\
 & LogicRAG & \textbf{56.41} & 45.16 & 46.95 & 21.53 & 38.64 & 9.42 & 13.19 & 9.39 \\
 & RT-RAG & 56.31 & 45.02 & 46.87 & 16.47 & \textbf{39.74} & 14.72 & 17.78 & 10.20 \\
 & ToQ & 53.37 & 47.05 & 47.60 & 32.44 & 39.47 & 22.31 & 24.77 & 11.79 \\
 & APT-RAG (Ours) & 56.04 & \textbf{52.39} & \textbf{52.15} & \textbf{52.18} & 36.92 & \textbf{24.60} & \textbf{25.83} & \textbf{22.58} \\
\bottomrule
\end{tabular}%
}
\caption{Performance comparison on the 300-example subset of MoNaCo and QAMPARI used to evaluate RT-RAG. Ans.\ P/R/F1 denote answer precision, recall, and F1; Ret.\ R denotes retrieval recall. Best values per metric, backbone, and benchmark are in \textbf{bold}.}
\label{tab:appendix_subset_result}
\end{table*}

\subsection{Adaptive Planning}
\label{app:prompts_planning}

Adaptive planning runs two prompts in sequence (\S\ref{sec:planning}): it first checks answerability, then performs decomposition if external evidence is required.

\paragraph{Answerability Checker.}
The check uses two main criteria: \textit{factual sufficiency}(are the entity and attributes explicit in the context?) and \textit{operational feasibility}(does the question only compare or aggregate given facts?). The prompt(Figure~\ref{fig:prompt_answerability}) produces a four-step trace and ends in a Boolean.
This split prevents the common conflation between "relevant information appears in the context" and "the question is fully answerable", which would otherwise trigger redundant retrieval on simple aggregations over sibling values.

\paragraph{Decomposer.}
A \textit{maintain}/\textit{split} decision is produced along with new sub-questions (Figure~\ref{fig:prompt_decomposer}). Mutual dependencies between these sub-questions are written as the placeholder $<Q_{j}>$ and are resolved during contextualization.
The prompt does not use a fixed depth or branching factor. Instead, it explicitly lists three conditions that prompt a split: when multiple bundled facts are present, when a stepwise dependency exists, or when operations such as sorting, ranking, or comparison are needed. This way, the tree's granularity reflects the input's structure rather than being governed by a global hyperparameter.

\subsection{Evidence Gathering}
\label{app:prompts_gathering}

External gathering is the only mode that issues a new retrieval call and therefore requires a dedicated query.
We isolate query rewriting from sub-question generation so the search query can be tuned for the retriever while the sub-question retains the verbosity required for downstream reasoning.

\paragraph{Single-Query Rewriting.}
A single sub-question is compressed into a compact keyword query for dense retrieval (Figure~\ref{fig:prompt_single_query_rewriter}).
Two short in-context examples specify the intended transformation — removing filler while preserving key entities and attributes.
This single sub-question rewriting path is used only when retrieval is dispatched from a non-decomposed question, typically the root. After decomposition occurs—when a node splits into siblings—the query rewriting process shifts to the batched variant described below.

\paragraph{Multi-Query Rewriting.}
In the batched rewriting process, a group of sibling sub-questions is rewritten in a single call (Figure~\ref{fig:prompt_multi_query_rewriter}), returning a JSON dictionary keyed by question index.
The instructions match those for single queries. Only the input and output formats change, ensuring rewriting stays consistent across batches.
The batch is formed at planning time: siblings routed to external gathering are rewritten together. As a result, the rewriting cost scales with the number of decomposition levels rather than the number of sibling nodes.

\subsection{Answer Generation}
\label{app:prompts_answering}
Each node produces an answer using a dedicated prompt for each gathering mode: lateral, external, and vertical(Figures~\ref{fig:prompt_lateral_answer_generator},~\ref{fig:prompt_external_answer_generator},~\ref{fig:prompt_vertical_answer_generator}). There is a unique prompt at the root(Figure~\ref{fig:prompt_final_answer_generator}), and a batched version for EC(Figure~\ref{fig:prompt_evidence_cluster}).
The prompts share the same system instruction and differ only in their input fields and integration rules.

\paragraph{Lateral Answer Generation.}
The input consists of the question and the preceding sibling QA pairs (Figure~\ref{fig:prompt_lateral_answer_generator}).
The prompt instructs the model to use only relevant QA pairs from siblings. 
One template handles both cases: when only one sibling is useful, and when many siblings are present but only a few are helpful. 
The output must be concise and self-contained so that the parent can use it as evidence for downstream reasoning.

\paragraph{External Answer Generation.}
Three inputs are required (Figure~\ref{fig:prompt_external_answer_generator}): the question, the retrieved documents, and the search query that produced them.
Including the search query helps the model focus on the retrieval intent.

\paragraph{Vertical Answer Generation.}
Child QA pairs are integrated into a single parent answer (Figure~\ref{fig:prompt_vertical_answer_generator}).
The prompt prohibits bullet-point or list-style output, since downstream generators consume the answer as a single evidence field.

\paragraph{Final Answer Generation.}
The root's child QA pairs are consolidated into the benchmark-formatted answer (Figure~\ref{fig:prompt_final_answer_generator}).
The system instruction comes from the MoNaCo answer-generation prompt. 
It includes the rule: "prefer fewer, correct answers over a long list that mixes correct and incorrect items." This instruction is used for both benchmarks.
We append a benchmark-conditional output block that retains MoNaCo's native answer-with-referent format and specifies a comma-separated list of short entity strings for QAMPARI.

\paragraph{Batched Answer Generation in EC.}
Answers for all sub-questions of an EC cluster are produced within a single LLM call (Figure~\ref{fig:prompt_evidence_cluster}; \S\ref{sec:evidence_aware_answer_batching}).
Each question is linked to its allowed documents. 
The prompt tells the model to use only those documents. Without this rule, the model might cross-reference documents between questions and confuse entities that share attributes.
The output is a JSON dictionary keyed by question index, so the cluster's answers can be redistributed to their corresponding nodes without ambiguity.

% =====================================================================
% G. Subset Evaluation against RT-RAG
% =====================================================================

\section{Subset Evaluation}
\label{app:subset}
We compare \FrameworkName{} against RT-RAG on a 300-example subset randomly sampled from each benchmark.
Table~\ref{tab:appendix_rt_cost} details inference cost, while Table~\ref{tab:appendix_subset_result} gives answer accuracy and retrieval coverage.

\paragraph{Inference Cost.}
We restrict the comparison to a subset because RT-RAG incurs substantially higher costs than the other baselines.
On MoNaCo with the 30B backbone, RT-RAG requires 115.6 LLM calls and consumes 1.13M input tokens per question, which is approximately $3.6\times$ more LLM calls and $8.4\times$ more input tokens compared to \FrameworkName{}.
This cost difference between RT-RAG and \FrameworkName{} is consistent across model scales and benchmarks.
This overhead stems from RT-RAG's repeated tree construction, consensus sampling, and sub-question rewriting, as detailed in Appendix~\ref{app:baselines_methods}.

% LDfQ 1, KnUc 5 - Main table
\begin{table*}[t]
\centering
\small
\setlength{\tabcolsep}{3.2pt}
\resizebox{\textwidth}{!}{%
\begin{tabular}{clrrrrrrrr}
\toprule
\multirow{2}{*}{Models} & \multicolumn{1}{c}{\multirow{2}{*}{Methods}} & \multicolumn{4}{c}{MoNaCo} & \multicolumn{4}{c}{QAMPARI} \\
\cmidrule(lr){3-6} \cmidrule(lr){7-10}
 &  & \# LLM Calls & Input Tok. & Output Tok. & Latency (s) & \# LLM Calls & Input Tok. & Output Tok. & Latency (s) \\
\midrule
\multirow{6}{*}{\textsc{Qwen3-4B-Inst}} & LLM-Only & 1.00 & 122 & 138 & 1.79 & 1.00 & 167 & 383 & 5.07 \\
 & NaiveRAG & 1.00 & 10,246 & 81 & 1.92 & 1.00 & 3,135 & 45 & 0.81 \\
 & Plan$^\ast$RAG & 5.25 & 28,759 & 499 & 27.51 & 2.55 & 4,028 & 168 & 2.48 \\
 & LogicRAG & 5.76 & 21,656 & 803 & 12.83 & 5.08 & 6,435 & 688 & 9.67 \\
 & \toq & 28.11 & 40,063 & 1,332 & 71.69 & 9.40 & 7,391 & 522 & 7.19 \\
 & \FrameworkName{} (Ours) & 14.96 & 62,515 & 1,630 & 29.62 & 4.92 & 8,252 & 229 & 3.60 \\
\midrule
\multirow{6}{*}{\textsc{Qwen3-30B-Inst}} & LLM-Only & 1.00 & 122 & 71 & 1.45 & 1.00 & 167 & 28 & 0.64 \\
 & NaiveRAG & 1.00 & 10,246 & 49 & 1.90 & 1.00 & 3,135 & 40 & 1.08 \\
 & Plan$^\ast$RAG & 7.66 & 42,538 & 875 & 22.20 & 5.25 & 10,172 & 398 & 8.69 \\
 & LogicRAG & 5.52 & 18,348 & 828 & 17.77 & 5.20 & 6,302 & 770 & 15.38 \\
 & \toq & 8.20 & 18,428 & 329 & 81.27 & 6.25 & 5,710 & 208 & 4.89 \\
 & \FrameworkName{} (Ours) & 33.61 & 138,920 & 4,596 & 104.59 & 11.69 & 21,898 & 1,153 & 24.29 \\
\bottomrule
\end{tabular}%
}
\caption{Per-question inference cost on the full MoNaCo and QAMPARI benchmarks.}
\label{tab:full_cost}
\end{table*}

\paragraph{Answer Quality.}
Even with this increased cost, RT-RAG does not consistently outperform \FrameworkName{}.
\FrameworkName{} leads in Answer F1 across most settings; the only exception is the 4B backbone on MoNaCo, where RT-RAG slightly surpasses \FrameworkName{} in F1 but achieves only about half of \FrameworkName{}'s Retrieval Recall—a gap that widens to nearly $3\times$ on MoNaCo with the 30B backbone.
We attribute this asymmetry to RT-RAG's iterative question refinement, which narrows the search space at the cost of evidence coverage.
Notably, scaling the backbone from 4B to 30B improves \FrameworkName{}'s MoNaCo F1 by $+13.0$, whereas RT-RAG's F1 improves by only $+7.0$ under the same conditions.
This result suggests that, for evidence-intensive QA, enhancement from a stronger backbone is more effectively achieved through broader evidence coverage—as in \FrameworkName{}—than through the narrower refinement approach used by RT-RAG.

\begin{center}
    \includegraphics[width=\linewidth]{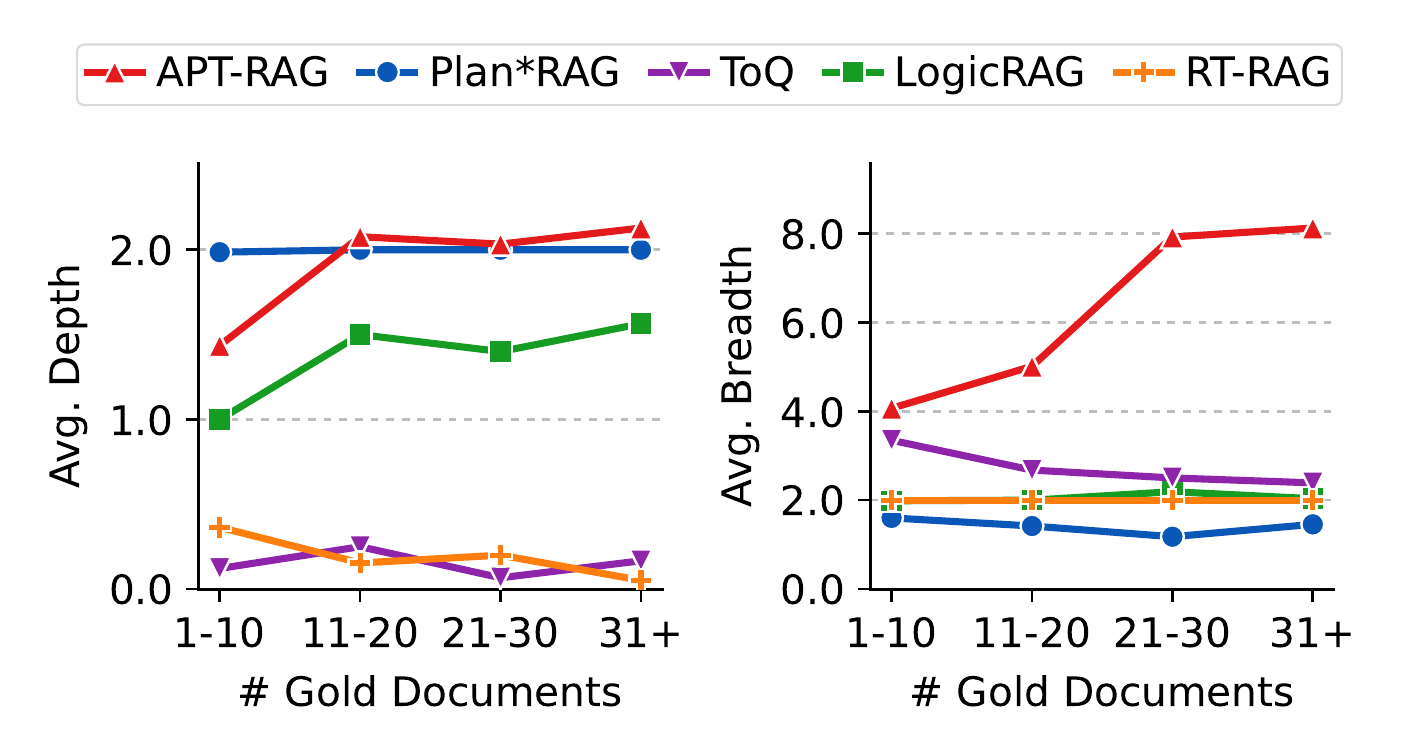}
    \captionof{figure}{
    Average depth (left) and breadth (right) on the 300-example subset, 
    grouped by the number of required gold documents.
    }
    \label{fig:appendix_depth_breadth}
\end{center}

\paragraph{Structural Adaptation.}
Figure~\ref{fig:appendix_depth_breadth} expands the comparison of depth and breadth metrics from Figure~\ref{fig:analysis_depth_breadth_adaptation}(b). It includes all baselines on the same subset.
For each example, tree depth is the maximum root-to-node distance in the constructed reasoning structure, and tree breadth is the average number of generated child sub-questions over decomposed parent nodes.
While the tree depth and breadth of \FrameworkName{} increase with the required evidence volume, those of the baselines remain nearly flat.
This difference reflects how each baseline builds its reasoning structure (see Appendix~\ref{app:baselines_methods}): \pstar and LogicRAG fix the graph before retrieval. RT-RAG builds a binary tree with limited depth. \toq stops expansion early when sufficiency checks are met.
Because of this rigid structure, each query gets limited reasoning space. This leads to the retrieval coverage gap seen in Table~\ref{tab:appendix_subset_result}.

% =====================================================================
% H. Full Evaluation
% =====================================================================
% LDfQ 1, KnUc 5 - Inference cost
% KnUc 2 - Statistical significance and uncertainty
\begin{table*}[t]
\centering
\small
\setlength{\tabcolsep}{3.2pt}
\resizebox{\textwidth}{!}{%
\begin{tabular}{clcccc}
\toprule
\multirow{2}{*}{Models} & \multicolumn{1}{c}{\multirow{2}{*}{Methods}} & \multicolumn{2}{c}{MoNaCo} & \multicolumn{2}{c}{QAMPARI} \\
\cmidrule(lr){3-4} \cmidrule(lr){5-6}
 &  & Ans.\ F1 & $\Delta$ Ans.\ F1 (95\% CI) & Ans.\ F1 & $\Delta$ Ans.\ F1 (95\% CI) \\
\midrule
\multirow{6}{*}{\textsc{Qwen3-4B-Inst}} & LLM-Only & 28.08 & 12.61 [10.37, 14.86] & 4.97 & 15.94 [14.45, 17.48] \\
 & NaiveRAG & 34.98 & 5.71 [3.82, 7.66] & 18.58 & 2.33 [1.37, 3.30] \\
 & Plan$^\ast$RAG & 27.02 & 13.67 [11.55, 15.82] & 5.40 & 15.51 [13.95, 17.03] \\
 & LogicRAG & 35.75 & 4.94 [2.92, 7.02] & 9.87 & 11.04 [9.72, 12.41] \\
 & \toq & 36.32 & 4.37 [2.46, 6.28] & 18.76 & 2.15 [1.33, 3.00] \\
 & \FrameworkName{} (Ours) & 40.69 & -- & 20.91 & -- \\
\midrule
\multirow{6}{*}{\textsc{Qwen3-30B-Inst}} & LLM-Only & 41.82 & 9.02 [6.71, 11.32] & 9.40 & 13.87 [12.35, 15.42] \\
 & NaiveRAG & 46.64 & 4.20 [2.06, 6.28] & 20.46 & 2.82 [1.63, 4.02] \\
 & Plan$^\ast$RAG & 47.03 & 3.81 [1.83, 5.85] & 17.67 & 5.60 [4.23, 6.98] \\
 & LogicRAG & 46.41 & 4.43 [2.38, 6.52] & 11.23 & 12.05 [10.62, 13.49] \\
 & \toq & 46.66 & 4.18 [2.11, 6.24] & 21.37 & 1.90 [0.81, 3.00] \\
 & \FrameworkName{} (Ours) & 50.84 & -- & 23.28 & -- \\
\bottomrule
\end{tabular}%
}
\caption{Answer F1 differences between \ours and each baseline on the full benchmarks ($\Delta$ Answer F1 $=$ \ours $-$ baseline), with 95\% confidence intervals.}
\label{tab:appendix_significance}
\end{table*}

\section{Full Evaluation}
\label{app:full_eval}

\paragraph{Inference Cost.}
Table~\ref{tab:full_cost} shows that \ours's inference cost depends on the evidence a question requires, rather than on a fixed budget.
On \monaco with the 30B backbone, where each question needs 43.3 gold pages on average, \ours is the most expensive method: it issues 33.61 LLM calls and takes 104.59s per question, while the structured RAG baselines issue 5.52 to 8.20 calls and take 17.77s to 81.27s.
On \qampari, where each question needs 13.0 pages, \ours's LLM calls drop by $2.9\times$ and its latency by $4.3\times$, while the baselines' calls drop by at most $1.5\times$.
With the smaller 4B backbone on QAMPARI, \ours is faster than LogicRAG and \toq, at 3.60s versus 9.67s and 7.19s.
Therefore, the baselines are less costly on MoNaCo not because of stricter budget constraints, but because the fixed structures discussed in \S\ref{sec:analysis} do not scale with increasing evidence requirements.

% =====================================================================
% I. Statistical Significance
% =====================================================================
% KnUc 2 - Statistical significance and uncertainty
\section{Statistical Significance}
\label{app:significance}

An aggregate F1 gap does not distinguish systematic advantage from within-set variation.
We therefore test each comparison for every question using a two-sided paired permutation test on the average F1 difference per question, with a 95\% confidence interval based on percentile bootstrap with 10,000 resamples.
All confidence intervals in Table~\ref{tab:appendix_significance} exclude zero, and every comparison yields $p \le 0.001$; all results remain significant after Bonferroni correction.
Consequently, the improvements reported in Table~\ref{tab:main_results} are statistically significant and not attributable to evaluation-set variation.

% =====================================================================
% J. Additional Ablation Results

\vspace{14 pt}
\begin{center}
    \small
    \setlength{\tabcolsep}{4pt}
    \begin{tabular}{lc}
    \toprule
    Method & Fine-grained Ret. R $\uparrow$ \\
    \midrule
    Fixed Breadth: 2 & 8.37 \\
    Fixed Breadth: Avg. & \underline{10.89} \\
    APT-RAG (Ours) & \textbf{11.22} \\
    \bottomrule
    \end{tabular}

    \captionof{table}{
    Fine-grained retrieval recall for the adaptive expansion ablation.
    }
    \label{tab:appendix_ablation_finegrained_retrieval}
\end{center}

\section{Additional Ablation Results}
\label{app:additional_ablation}

Table~\ref{tab:appendix_ablation_finegrained_retrieval} reports the adaptive expansion ablation under MoNaCo's fine-grained evidence matching criterion.
MoNaCo gold evidence is annotated not only with the document title or URL, but also with the specific section within the document and the content type of the evidence, such as sentence, list, or table. 
Since our corpus is constructed using these fine-grained evidence units, this analysis evaluates whether each method retrieves evidence at the same granularity.
Under this stricter criterion, \FrameworkName{} achieves the highest retrieval recall.
Together with the higher Answer F1 in Table~\ref{tab:component_ablation}, this result suggests that adaptive expansion improves answer quality partly by recovering more complete evidence at the corpus construction granularity.

\begin{center}
    \small
    \setlength{\tabcolsep}{2pt}
    \begin{tabular}{lcccccc}
    \toprule
    & \multicolumn{3}{c}{MoNaCo} & \multicolumn{3}{c}{QAMPARI} \\
    \cmidrule(lr){2-4} \cmidrule(lr){5-7}
    Models & LG & VG & EG & LG & VG & EG \\
    \midrule
    Qwen3-4B-Inst
        & \textbf{4.0\%} & 17.3\% & 78.7\%
        & \textbf{1.6\%} & 12.0\% & 86.4\% \\
    Qwen3-30B-Inst
        & \textbf{6.2\%} & 16.6\% & 77.2\%
        & \textbf{3.7\%} & 19.8\% & 76.5\% \\
    \bottomrule
    \end{tabular}
    
    \captionof{table}{
        Usage statistics for topology-aware evidence gathering.
    }
    \label{tab:topology_aware_gathering_usage}
\end{center}

\begin{table}[!b]
\centering
\small
\setlength{\tabcolsep}{4pt}
\begin{tabular}{llcc}
\toprule
\textbf{Model} & \textbf{Method}
& \shortstack{\textbf{Gemini-3.5-}\\\textbf{Flash}}
& \textbf{GPT-5.4} \\
\midrule
\multirow{6}{*}{Qwen3-4B}
& LLM-Only       & 27.08 & 28.08 \\
& NaiveRAG       & 34.19 & 34.98 \\
& PlanRAG        & 26.40 & 27.02 \\
& LogicRAG       & 35.31 & 35.75 \\
& ToQ            & 35.39 & 36.32 \\
& APT-RAG (Ours) & \textbf{40.04} & \textbf{40.69} \\
\midrule
\multirow{6}{*}{Qwen3-30B}
& LLM-Only       & 41.16 & 41.82 \\
& NaiveRAG       & 45.49 & 46.64 \\
& PlanRAG        & 46.60 & 47.03 \\
& LogicRAG       & 46.16 & 46.41 \\
& ToQ            & 45.91 & 46.66 \\
& APT-RAG (Ours) & \textbf{50.15} & \textbf{50.84} \\
\bottomrule
\end{tabular}

\caption{
Answer F1 on \monaco using Gemini-3.5-Flash and GPT-5.4 as LLM-as-a-judge evaluators.
}
\label{tab:evaluator_comparison}
\end{table}

\paragraph{Topology-Aware Gathering Usage.}
Table~\ref{tab:topology_aware_gathering_usage} reports the proportion of nodes resolved by each evidence-gathering strategy across datasets and model scales. 
Lateral gathering (LG) accounts for a small fraction (1.6-6.2\%) of the total nodes generated, while most nodes rely on external or vertical gathering.
Thus, the majority of sibling nodes remain independent and can be processed in parallel.
Despite its limited usage, LG substantially reduces latency by avoiding redundant retrieval and tree expansion, as demonstrated by the ablation in Figure~\ref{fig:ablation_latency_distribution}.

\paragraph{Evaluator Robustness.}
\label{app:evaluator_robustness}
To assess the robustness of the LLM-as-a-judge evaluation on \monaco, we additionally evaluate all methods using Gemini-3.5-Flash as an independent evaluator. As shown in Table~\ref{tab:evaluator_comparison}, the two evaluators yield comparable F1 scores across all methods and backbone scales. Importantly, APT-RAG consistently achieves the highest F1 under both evaluators, indicating that our main performance conclusions are robust to the choice of evaluator.

% =====================================================================
% F. Case Study
% =====================================================================
\section{Case Study}
\label{app:case_study}

Figure~\ref{fig:case_study_mexico} presents a qualitative inference trace from MoNaCo, illustrating how \FrameworkName{} operates through adaptive planning and topology-aware evidence gathering.
This is the same question used in Figure~\ref{fig:structured_rag_limit}, where we illustrate the limitations of existing structured RAG methods.
The example requires identifying Mexico's wars whose soldier counts are greater than 100{,}000 and fewer than 500{,}000.

% =====================================================================
% Figure
% =====================================================================
\clearpage
% ============================================================
% Case-study formatting commands
% ============================================================

% #1: question index, e.g., $q_1$
% #2: question text
\newtcolorbox{tracenode}[2]{
    enhanced,
    colback=black!1,
    colframe=black!55,
    colbacktitle=black!7,
    coltitle=black,
    boxrule=0.4pt,
    arc=0.8mm,
    left=1.5mm,
    right=1.5mm,
    top=1.0mm,
    bottom=1.0mm,
    before skip=1.5mm,
    after skip=1.5mm,
    fonttitle=\small,
    title={\textbf{Sub-question #1:} #2}
}

% #1: field label without punctuation
% #2: field content
\newcommand{\tracefield}[2]{%
    \textcolor{black!65}{\textbf{#1:}}\enspace
    #2\par
}

\newcommand{\answerable}{%
    \textcolor{green!45!black}{\textbf{Answerable}}%
}

\newcommand{\unanswerable}{%
    \textcolor{orange!80!black}{\textbf{Unanswerable}}%
}

% ============================================================
% Page 1
% ============================================================

\begin{figure*}[t]
\centering

\begin{tcolorbox}[
    enhanced,
    colback=white,
    colframe=black!80,
    colbacktitle=black!60,
    coltitle=white,
    boxrule=0.6pt,
    arc=1mm,
    left=1.0mm,
    right=1.0mm,
    top=1.0mm,
    bottom=1.0mm,
    boxsep=1.1mm,
    fonttitle=\bfseries,
    title={
        Inference Case: Evidence-Intensive Counting over Mexico's Wars
    }
]

\small
\setlength{\parindent}{0pt}
\setlength{\parskip}{0pt}
\raggedright

% ------------------------------------------------------------
% Initial question and final answer
% ------------------------------------------------------------

\begin{tcolorbox}[
    enhanced,
    colback=blue!3,
    colframe=blue!35!black,
    boxrule=0.45pt,
    arc=0.8mm,
    left=1.5mm,
    right=1.5mm,
    top=1.0mm,
    bottom=1.0mm
]

\textbf{Question $\ques$:}
How many of Mexico's wars were fought with more than
100{,}000 soldiers but fewer than 500{,}000?

\smallskip

\tracefield{Evidence}{
    $q_1,a_1;\ \ldots;\ q_4,a_4$
}

\tracefield{Final answer $\ans$}{
    \textcolor{green!45!black}{\textbf{3 \checkmark}}
}

\end{tcolorbox}

\medskip
\textbf{Inference trace:}

% ------------------------------------------------------------
% q_1
% ------------------------------------------------------------

\begin{tracenode}
    {$q_1$}
    {What are the wars fought by Mexico throughout its history?}

\tracefield{Evidence}{
    $q_{1,1},a_{1,1};\ q_{1,2},a_{1,2}$
}

\tracefield{Sub-answer $a_1$}{
    Mexico's wars span independence, territorial, intervention,
    revolutionary, religious, global-war, and internal conflicts,
    including the Mexican War of Independence, $\ldots$, and the
    Dirty War.
}

\end{tracenode}

% ------------------------------------------------------------
% q_1,1
% ------------------------------------------------------------

\begin{tracenode}
    {$q_{1,1}$}
    {What are the major wars fought by Mexico throughout its history?}

\tracefield{Evidence}{
    $\doci{1}$: Mexican War of Independence, $\ldots$,
    $\doci{20}$: Dirty War, $\ldots$
}

\tracefield{Sub-answer $a_{1,1}$}{
    The major wars include the Mexican War of Independence,
    Texas--Indian Wars, Mexican--American War, French Intervention
    in Mexico, Mexican Revolution, Cristero War, World War II
    involvement, $\ldots$, and the Dirty War.
}

\end{tracenode}

% ------------------------------------------------------------
% q_1,2
% ------------------------------------------------------------

\begin{tracenode}
    {$q_{1,2}$}
    {For each war in \texttt{<$q_{1,1}$>}, provide a brief
    description of its causes, key events, and outcomes.}

\tracefield{Contextualized question}{
    For each of the following wars fought by Mexico throughout its
    history---Mexican War of Independence, $\ldots$, and the Dirty
    War---provide a brief description of its causes, key events,
    and outcomes.
}

\tracefield{Answerability}{
    \unanswerable
}

\tracefield{Evidence}{
    $q_{1,2,1},a_{1,2,1};\ \ldots;\
    q_{1,2,10},a_{1,2,10}$
}

\tracefield{Sub-answer $a_{1,2}$}{
    The wars are described by their main causes and outcomes:
    independence from Spain, Indigenous displacement, territorial
    loss, French intervention, constitutional reform,
    church--state conflict, World War II mobilization, $\ldots$
}

\end{tracenode}

% ------------------------------------------------------------
% q_1,2,1
% ------------------------------------------------------------

\begin{tracenode}
    {$q_{1,2,1}$}
    {What were the causes, key events, and outcomes of the
    Mexican War of Independence (1810--1821)?}

\tracefield{Evidence}{
    $\doci{1}$: Mexican War of Independence, $\ldots$, Cry of
    Dolores, Miguel Hidalgo, Army of the Three Guarantees,
    $\ldots$; $\doci{20}$: Treaty of Cordoba, casualties,
    $\ldots$
}

\tracefield{Sub-answer $a_{1,2,1}$}{
    Napoleon's invasion of Spain created a legitimacy crisis;
    the Cry of Dolores, Hidalgo, Morelos, Plan of Iguala, and
    Treaty of Cordoba, $\ldots$, led to independence.
}

\end{tracenode}

% ------------------------------------------------------------
% q_1,2,2
% ------------------------------------------------------------

\begin{tracenode}
    {$q_{1,2,2}$}
    {What were the causes, key events, and outcomes of the
    Texas--Indian Wars (1820--1875)?}

\tracefield{Evidence}{
    $\doci{1}$: Texas--Indian Wars, $\ldots$, Comanche, Apache,
    Anglo-American settlement, $\ldots$;
    $\doci{20}$: Texas Revolution, Comanche--Mexico Wars,
    $\ldots$
}

\tracefield{Sub-answer $a_{1,2,2}$}{
    Anglo-American settlement conflicted with Comanche and Apache
    sovereignty; raids, the Great Raid of 1840, $\ldots$, ended in
    Indigenous displacement.
}

\end{tracenode}

\begin{center}
    \textcolor{black!55}{$\vdots$}
\end{center}

% ------------------------------------------------------------
% q_1,2,10
% ------------------------------------------------------------

\begin{tracenode}
    {$q_{1,2,10}$}
    {What were the causes, key events, and outcomes of the Dirty
    War in Mexico (1960s--1980s)?}

\tracefield{Evidence}{
    $\doci{1}$: Mexican Dirty War, $\ldots$, PRI, state repression,
    left-wing groups, disappearances, $\ldots$;
    $\doci{20}$: Tlatelolco, Corpus Christi, guerrilla groups,
    $\ldots$
}

\tracefield{Sub-answer $a_{1,2,10}$}{
    Under PRI rule, the Dirty War suppressed left-wing students
    and guerrillas through Tlatelolco, Corpus Christi, torture,
    disappearances, and extrajudicial violence.
}

\end{tracenode}

\end{tcolorbox}
\end{figure*}

\clearpage

% ============================================================
% Page 2
% ============================================================

\begin{figure*}[t]
\centering

\begin{tcolorbox}[
    enhanced,
    colback=white,
    colframe=black!80,
    colbacktitle=black!60,
    coltitle=white,
    boxrule=0.6pt,
    arc=1mm,
    left=1.0mm,
    right=1.0mm,
    top=1.0mm,
    bottom=1.0mm,
    boxsep=1.1mm,
    fonttitle=\bfseries,
    title={Inference Case: Continued}
]

\small
\setlength{\parindent}{0pt}
\setlength{\parskip}{0pt}
\raggedright

% ------------------------------------------------------------
% q_2
% ------------------------------------------------------------

\begin{tracenode}
    {$q_2$}
    {For each war in \texttt{<$q_1$>}, what was the number of
    soldiers involved?}

\tracefield{Contextualized question}{
    For each war throughout Mexico's history, what was the number
    of soldiers involved: Mexican War of Independence
    (1810--1821), $\ldots$, Dirty War (1960s--1980s)?
}

\tracefield{Answerability}{
    \unanswerable
}

\tracefield{Evidence}{
    $q_{2,1},a_{2,1};\ \ldots;\ q_{2,10},a_{2,10}$
}

\tracefield{Sub-answer $a_2$}{
    For the Mexican War of Independence (1810--1821), exact troop
    counts are not stated, but casualties are estimated at
    250{,}000--600{,}000; $\ldots$; for the Dirty War
    (1960s--1980s), the evidence reports about 784{,}300 personnel.
}

\end{tracenode}

% ------------------------------------------------------------
% q_2,1
% ------------------------------------------------------------

\begin{tracenode}
    {$q_{2,1}$}
    {What was the number of soldiers involved in the Mexican War
    of Independence (1810--1821)?}

\tracefield{Evidence}{
    $\doci{1}$: Mexican War of Independence, $\ldots$, insurgent
    forces, Spanish Empire army, casualties
    250{,}000--600{,}000, $\ldots$;
    $\doci{20}$: Treaty of Cordoba, Plan of Iguala, $\ldots$
}

\tracefield{Sub-answer $a_{2,1}$}{
    Exact troop counts are not given; the evidence mentions
    insurgent forces, the Spanish Empire's army, and casualties
    around 250{,}000--600{,}000.
}

\end{tracenode}

% ------------------------------------------------------------
% q_2,2
% ------------------------------------------------------------

\begin{tracenode}
    {$q_{2,2}$}
    {What was the number of soldiers involved in the
    Texas--Indian Wars (1820--1875)?}

\tracefield{Evidence}{
    $\doci{1}$: Texas--Indian Wars, $\ldots$, Comanche, Apache,
    Republic of Texas, U.S. Army, $\ldots$;
    $\doci{20}$: Texas Revolution, Apache--Mexico Wars, $\ldots$
}

\tracefield{Sub-answer $a_{2,2}$}{
    No single total is given; the evidence mentions the Republic
    of Texas, U.S. Army, Comanche, Kiowa, Apache, $\ldots$, across
    many engagements.
}

\end{tracenode}

\begin{center}
    \textcolor{black!55}{$\vdots$}
\end{center}

% ------------------------------------------------------------
% q_2,10
% ------------------------------------------------------------

\begin{tracenode}
    {$q_{2,10}$}
    {What was the number of soldiers involved in the Dirty War
    (1960s--1980s)?}

\tracefield{Evidence}{
    $\doci{1}$: Mexican Dirty War, $\ldots$, military, police,
    National Guard, $\ldots$;
    $\doci{20}$: Mexican drug war, $\ldots$, 277{,}000 soldiers,
    368{,}000 police, and 107{,}000 National Guard personnel,
    $\ldots$
}

\tracefield{Sub-answer $a_{2,10}$}{
    The evidence mentions about 784{,}300 total security personnel,
    including 277{,}000 soldiers, 368{,}000 police officers, and
    107{,}000 National Guard personnel.
}

\end{tracenode}

% ------------------------------------------------------------
% q_3
% ------------------------------------------------------------

\begin{tracenode}
    {$q_3$}
    {From the wars in \texttt{<$q_2$>}, which ones had more than
    100{,}000 soldiers but fewer than 500{,}000 soldiers?}

\tracefield{Contextualized question}{
    From the wars in the following war-count evidence, which ones
    had more than 100{,}000 soldiers but fewer than 500{,}000
    soldiers: Mexican War of Independence
    (250{,}000--600{,}000 casualties), $\ldots$, Dirty War
    (784{,}300 personnel)?
}

\tracefield{Answerability}{
    \answerable
}

\tracefield{Evidence}{
    $q_2,a_2$
}

\tracefield{Sub-answer $a_3$}{
    Mexican War of Independence, Reform War, and Cristero War
    satisfy the condition. Wars with missing totals remain
    uncertain, smaller rebellions fall below 100{,}000, and the
    Dirty War exceeds the upper bound.
}

\end{tracenode}

% ------------------------------------------------------------
% q_4
% ------------------------------------------------------------

\begin{tracenode}
    {$q_4$}
    {How many wars satisfy the condition in the final candidate set?}

\tracefield{Contextualized question}{
    How many wars satisfy the condition among the final candidate
    set: Mexican War of Independence, Reform War, and Cristero War?
}

\tracefield{Answerability}{
    \answerable
}

\tracefield{Evidence}{
    $q_3,a_3$
}

\tracefield{Sub-answer $a_4$}{
    \textcolor{green!45!black}{\textbf{3}}
}

\end{tracenode}

\end{tcolorbox}

\caption{Inference case for the MoNaCo dataset.}
\label{fig:case_study_mexico}
\end{figure*}

\clearpage

\begin{figure*}[p]
\begin{tcblisting}{appprompt, title={Prompt template for Contextualization}}
You are a Contextualizer who extracts the key keywords needed to specify a vague query and rewrites the question in a more specific form.

[Input Information]
- Target question: A question that requires specification. This question may contain reference expressions such as <Qn>.
  - <Qn> refers to the result (answer) of the nth question.
- Prior QA information: Previous questions and their corresponding answers that are necessary to specify the target question.

[Process]
- Step 1. Analyze the target question.
- Step 2. Identify required search conditions.
- Step 3. Extract complete structured items from prior answers.
- Step 4. Rewrite the question by replacing vague references with concrete information.

[Output Format]
{
  "step 1": "...",
  "step 2": "...",
  "step 3": ["keyword1", "keyword2", ...],
  "step 4": "Rewritten question"
}

Target question: {target_question}
Prior QA information: {prior_qa_information}
\end{tcblisting}
\caption{Prompt template for contextualizing question.}
\label{fig:prompt_contextualizer}
\end{figure*}

\begin{figure*}[p]
\begin{tcblisting}{appprompt, title={Prompt template for Answerability Check}}
You are an answerability checker who determines whether external search is required to answer the current question.

[Input]
- input question: The question for which you must determine whether search is necessary.
- context: Previous question-answer (QA) pairs that provide context relevant to the current question.

[Decision Rules]
- Skip search if the core entity and required attributes are already explicit in the context.
- Perform search if the target entity or required attributes are missing or uncertain.
- Skip search if the question only compares, summarizes, or reorganizes information already available in the context.
- Perform search if the question requires new information.

[Output Format]
{
  "step 1": "...",
  "step 2": "...",
  "step 3": "...",
  "step 4": true or false
}

input question: {input_question}
context: {context}
\end{tcblisting}
\caption{Prompt template for the Answerability Checker (\S\ref{sec:planning}).}
\label{fig:prompt_answerability}
\end{figure*}

\begin{figure*}[p]
\begin{tcblisting}{appprompt, title={Prompt template for Decomposition}}
You are a Decomposer who analyzes an input question and determines whether it should be handled as a single search (maintain) or decomposed into sub-queries (split).

[Maintain / Split Decision Criteria]
A question should be split when multiple pieces of data must be assembled, when the answer requires stepwise dependency, or when the task requires sorting, ranking, comparison, or other complex operations.
A question should be maintained when the requested information is likely to be found as a single bundled fact.

[Decomposition Rules]
- If a sub-query depends on the result of a previous query, indicate this dependency using <Qn>, where n refers to the required prior query.

[Output Format]
{
  "decision": "maintain" or "split",
  "subqueries": [] or ["sub-query 1", "sub-query 2", ...]
}

input question: {input_question}
\end{tcblisting}
\caption{Prompt template for the Decomposer (\S\ref{sec:planning}).}
\label{fig:prompt_decomposer}
\end{figure*}

\begin{figure*}[p]
\begin{tcblisting}{appprompt, title={Prompt template for Lateral Answer Generation}}
[System prompt]
You are a helpful question-answering assistant. Your task is to answer a complex question provided by the user.
You may generate a brief explanation before presenting the final answer if necessary.
Each sentence must be concise and clear. Since the answer will be passed to another agent for follow-up tasks, preserve sufficient context while remaining concise.
Your response must strictly follow the format below:
Answers: {ANSWERS}

[Input prompt]
Question: {question}

QA pairs generated along the path leading to the current question:
Q1: {previous_question_1}
A1: {previous_answer_1}
...

Above are related question-answer pairs generated along the path leading to the current question.
Some QA pairs may contain information relevant to answering the current question.
Use relevant QA pairs and ignore irrelevant ones.

Answer:
\end{tcblisting}
\caption{Prompt template for answer generation under lateral gathering (\S\ref{sec:evidence_gathering}).}
\label{fig:prompt_lateral_answer_generator}
\end{figure*}

\begin{figure*}[p]
\begin{tcblisting}{appprompt, title={Prompt template for External Answer Generation}}
[System prompt]
You are a helpful question-answering assistant. Your task is to answer a complex question provided by the user.
You may generate a brief explanation before presenting the final answer if necessary.
Each sentence must be concise and clear. Since the answer will be passed to another agent for follow-up tasks, preserve sufficient context while remaining concise.
Your response must strictly follow the format below:
Answers: {ANSWERS}

[Input prompt]
Question: {question}
Search Query: {query}
Documents: {documents}

Above are multiple excerpts of paragraphs and tables. Each document has a title followed by the actual content.
Some documents may contain helpful information for answering the question.
Use relevant documents and ignore irrelevant ones.

Answer:
\end{tcblisting}
\caption{Prompt template for answer generation under external gathering (\S\ref{sec:evidence_gathering}).}
\label{fig:prompt_external_answer_generator}
\end{figure*}
\begin{figure*}[p]
\begin{tcblisting}{appprompt, title={Prompt template for Vertical Answer Generation}}
[System prompt]
You are a helpful parent-node answer generator.
Your task is to integrate child sub-question answers into a single parent answer without losing meaningful information.
If the child QA pairs contain relevant details, include them in the final answer.
Provide a concise yet complete response, and do not use bullet points or numbered lists unless explicitly required.
Your response must strictly follow the format below:
Answers: {ANSWERS}

[Input prompt]
Current Question: {question}

sub-QA pairs for the current question:
Q1: {sub_question_1}
A1: {sub_answer_1}
...

Above are multiple sub question-answer pairs related to the current question.
Synthesize the provided sub QA pairs into the most complete and comprehensive answer to the current question.

Answer:
\end{tcblisting}
\caption{Prompt template for answer generation under vertical gathering (\S\ref{sec:evidence_gathering}).}
\label{fig:prompt_vertical_answer_generator}
\end{figure*}

\begin{figure*}[p]
\begin{tcblisting}{appprompt, title={Prompt template for Final Answer Generation}}
[System prompt]
You are a helpful question answering assistant. Your task is to answer the question with high precision.
- List only answers that directly and clearly satisfy what the question asks.
- Prefer fewer, correct answers over a long list that mixes correct and incorrect items. When in doubt, omit the item.

[For QAMPARI Only]
- Each answer must be a short answer (entity or short phrase), not a full sentence.
- If multiple answers exist, separate them using commas.
- List each distinct answer only once.
- Do not include any explanation or extra text before or after the answer line.
- Your response must use the following format:
Answers: answer1, answer2, ..., answerN

[For MoNaCo Only]
- Preserve relevant context so each part of the answer is clearly tied to what it refers to.
- Your response must use the following format:
Answers: {ANSWERS}

[Input prompt]
Current Question: {question}

sub-QA pairs for the current question:
Q1: {sub_question_1}
A1: {sub_answer_1}
...

Using the provided sub QA pairs, generate a concise and direct answer to the current question.
Include only what is needed to answer the question; omit filler and redundant phrasing.

Answer:
\end{tcblisting}
\caption{Prompt template for final answer generation at the root.}
\label{fig:prompt_final_answer_generator}
\end{figure*}

\begin{figure*}[p]
\begin{tcblisting}{appprompt, title={Prompt template for Single-Query Rewriting}}
- You are a model that rewrites users' questions into search engine queries.
- Rewrite the question into one optimal search query.
- Examples:
  Please recommend a national park in California with hiking trails similar to Yosemite.
  Query: California national park hiking trails similar to Yosemite
  Please find the admission fee for the Louvre Museum.
  Query: Louvre Museum admission fee

Question: {question}
Query:
\end{tcblisting}
\caption{Prompt template for single-query rewriting before external retrieval.}
\label{fig:prompt_single_query_rewriter}
\end{figure*}

\begin{figure*}[p]
\begin{tcblisting}{appprompt, title={Prompt template for Multi-Query Rewriting}}
- You are a model that rewrites users' questions into search engine queries.
- Rewrite each question into one optimal search query.
- Use the question numbers as JSON keys. Each value should contain only the rewritten search query corresponding to its question.

Question:
{questions}

Output format:
{
  "1": "query",
  "2": "query",
  ...
}

Query:
\end{tcblisting}
\caption{Prompt template for multi-query rewriting before external retrieval.}
\label{fig:prompt_multi_query_rewriter}
\end{figure*}

\begin{figure*}[p]
\begin{tcblisting}{appprompt, title={Prompt template for Batched Answer Generation}}
Each question includes an explicit list of associated documents. Use only the documents in that list for the question.

Rules:
- Answer each question independently.
- For each question, use only the documents specified in its list and no others.
- Use the question numbers as JSON keys. Each value should contain only the answer to its corresponding question.

Question-document ID pairs:
[Question 1]
Documents: [D1, D2, ...]
Question: {sub_question_1}
Search Query: {search_query_1}

[Question 2]
Documents: [D2, D3, ...]
Question: {sub_question_2}
Search Query: {search_query_2}

Documents:
{documents}

Output format:
{
  "1": "answer",
  "2": "answer",
  ...
}

Answer:
\end{tcblisting}
\caption{Prompt template for generating answers over sub-question clusters in Evidence-Guided Batched Answer Generation (see Section \ref{sec:evidence_aware_answer_batching}).}
\label{fig:prompt_evidence_cluster}
\end{figure*}

\end{document}